\documentclass{article}
\usepackage{ijcai20}

\usepackage{times}

\usepackage{amsmath}
\usepackage{amsfonts}
\usepackage{amssymb}
\usepackage{amsthm}
\usepackage{graphicx}
\usepackage{algpseudocode}
\usepackage{algorithm}
\usepackage{subcaption,graphicx}
\usepackage{url}
\usepackage[hidelinks]{hyperref}
\usepackage[utf8]{inputenc}
\usepackage{comment}
\usepackage{xcolor}
\usepackage{mathrsfs}
\usepackage{caption}
\DeclareCaptionFont{9pt}{\fontsize{9pt}{10pt}\selectfont}
\renewcommand{\footnotesize}{\fontsize{9pt}{11pt}\selectfont}

\usepackage{booktabs}
\newcommand{\ra}[1]{\renewcommand{\arraystretch}{#1}}

\title{Measuring the Discrepancy between Conditional Distributions:\\ Methods, Properties and Applications}

\author{
Shujian Yu$^1$\footnote{Contact Author}\and
Ammar Shaker$^1$\and
Francesco Alesiani$^{1}$\And
Jose Principe$^2$\\
\affiliations
$^1$NEC Labs Europe, $69115$ Heidelberg, Germany\\
$^2$University of Florida, Gainesville, FL $32611$, USA\\
\emails
\{Shujian.Yu, Ammar.Shaker, Francesco.Alesiani\}@neclab.eu,
principe@cnel.ufl.edu
}

\usepackage{mathtools}

\DeclarePairedDelimiter{\diagfences}{(}{)}
\newtheorem{theorem}{Theorem}
\newtheorem{lemma}[theorem]{Lemma}

\newcommand{\tr}{\operatorname{tr}\diagfences}

\DeclareMathOperator{\Tr}{tr}
\DeclareMathOperator{\range}{range}

\newtheorem{property}{Property}[section]
\newtheorem{corollary}{Corollary}[section]

\begin{document}
	
\maketitle

\begin{abstract}
We propose a simple yet powerful test statistic to quantify the discrepancy between two conditional distributions. The new statistic avoids the explicit estimation of the underlying distributions in high-dimensional space and it operates on the cone of symmetric positive semidefinite (SPS) matrix using the Bregman matrix divergence. Moreover, it inherits the merits of the correntropy function to explicitly incorporate high-order statistics in the data. We present the properties of our new statistic and illustrate its connections to prior art. We finally show the applications of our new statistic on three different machine learning problems, namely the multi-task learning over graphs, the concept drift detection, and the information-theoretic feature selection, to demonstrate its utility and advantage. Code of our statistic is available at \url{https://bit.ly/BregmanCorrentropy}.
\end{abstract}

\section{Introduction}

Measuring the discrepancy or divergence between two conditional distribution functions plays a leading role in numerous real-world machine learning problems. One vivid example is the modeling of the seasonal effects on consumer preferences, in which the statistical analyst needs to distinguish the changes on the distributions of the merchandise sales conditioning on the explanatory variables such as the amount of money spent on advertising, the promotions being run, etc.


Despite substantial efforts have been made on specifying the discrepancy for unconditional distribution (density) functions (see~\cite{anderson1994two,gretton2012kernel,pardo2018statistical} and the references therein), methods on quantifying the discrepancy of regression models or statistical tests for conditional distributions are scarce.



Prior art falls into two categories. The first relies heavily on the precise estimation of the underlying distribution functions using different density estimators, such as the $k$-Nearest Neighbor (kNN) estimator~\cite{wang2009divergence} and the kernel density estimator (KDE)~\cite{lee2006estimation}.
However, density estimation is notoriously difficult for high-dimensional data. Moreover, existing conditional tests (e.g.,~\cite{zheng2000consistent,fan2006nonparametric}) are always one-sample based, which means that they are designed to test if the observations are generated by a conditional distribution $p(y|x)$ in a particular parametric family with parameter $\theta$, rather than distinguishing $p_1(y|x)$ from $p_2(y|x)$.
Another category defines a distance metric through the embedding of probability measures in another space (typically the reproducing kernel Hilbert space or RKHS). A notable example is the Maximum Mean Discrepancy (MMD)~\cite{gretton2012kernel} which has attracted much attention in recent years due to its solid mathematical foundation. However, MMD always requires high computational burden and carefully hyper-parameter (e.g., the kernel width) tuning~\cite{gretton2012optimal}. Moreover, computing the distance of the embeddings of two conditional distributions in RKHS still remains a challenging problem~\cite{ren2016conditional}.

Different from previous efforts, we propose a simple statistic to quantify the discrepancy between two conditional distributions. It directly operates on the cone of symmetric positive semidefinite (SPS) matrix to avoid the estimation of the underlying distributions. To strengthen the discriminative power of our statistic, we make use of the correntropy function~\cite{santamaria2006generalized}, which has demonstrated its effectiveness in non-Gaussian signal processing~\cite{liu2007correntropy}, to explicitly incorporate higher order information in the data. We demonstrate the power of our statistic and establish its connections to prior art. Three solid examples of machine learning applications are presented to demonstrate the effectiveness and the superiority of our statistic.

\section{Background Knowledge}

\subsection{Bregman Matrix Divergence and Its Computation}


A symmetric matrix is positive semidefinite (SPS) if all its eigenvalues are non-negative. We denote $\mathcal{S}_{+}^n$ the set of all $n\times n$ SPS matrices, i.e., $\mathcal{S}_{+}^n=\{A\in \mathbb{R}^{n\times n}|A=A^T,A\succcurlyeq0\}$.
To measure the nearness between two SPS matrices, a reliable choice is the Bregman matrix divergence~\cite{kulis2009low}. Specifically, given a strictly convex, differentiable function $\varphi$ that maps matrices to the extended real numbers, the Bregman divergence from the matrix $\rho$ to the matrix $\sigma$ is defined as:
\begin{equation}
D_{\varphi,B}(\sigma\|\rho) = \varphi(\sigma) - \varphi(\rho) - \tr{(\nabla\varphi(\rho))^T(\sigma - \rho)},
\end{equation}
where $\tr A$ denotes the trace of matrix $A$.

When $\varphi(\sigma) = \Tr(\sigma\log\sigma - \sigma)$, where $\log\sigma$ is the matrix logarithm, the resulting Bregman divergence is:
\begin{eqnarray}\label{eq:vNRelativeEntropy}
D_{vN}(\sigma\|\rho) = \Tr(\sigma \log \sigma - \sigma \log \rho - \sigma + \rho),
\end{eqnarray}
which is also referred to von Neumann divergence in quantum information theory~\cite{nielsen2002quantum}. Another important matrix divergence arises by taking $\varphi(\sigma)=-\log\det \sigma$, in which the resulting Bregman divergence reduces to:
\begin{eqnarray}\label{eq:LogDet}
D_{\ell D}(\sigma\|\rho) = \Tr(\rho^{-1}\sigma) + \log_2\frac{|\rho|}{|\sigma|} - n,
\end{eqnarray}
and is commonly called the LogDet divergence.

\subsection{Correntropy Function: A Generalized Correlation Measure}
The correntropy function of two random variables $x$ and $y$ is defined as~\cite{santamaria2006generalized}:
\begin{equation}
V(x,y) = \mathbb{E}[\kappa(x,y)] = \iint \kappa(x,y)dF_{X,Y}(x,y),
\end{equation}
where $\mathbb{E}$ denotes mathematical expectation, $\kappa$ is a positive definite kernel function, and $F_{X,Y}(x,y)$ is the joint distribution of (X,Y). One widely used kernel function is the Gaussian kernel given by:
\begin{equation}
\kappa(x,y) = \frac{1}{\sqrt{2\pi}\sigma}\exp{\{-\frac{(x-y)^2}{2\sigma^2}\}}.
\end{equation}

Taking Taylor series expansion of the Gaussian kernel, we have:
\begin{equation}\label{eq:taylor_extension}
V_\sigma(x,y)=\frac{1}{\sqrt{2\pi}\sigma}\sum_{n=0}^\infty \frac{(-1)^n}{2^nn!}\mathbb{E}[\frac{(x-y)^{2n}}{\sigma^{2n}}].
\end{equation}


Therefore, correntropy involves all the even moments\footnote{A different kernel would yield a different expansion, for instance the sigmoid kernel $\kappa(x,y)=\tanh (\langle x,y\rangle+\theta)$ admits an expansion in terms of the odd moments of its argument.} of random variable $e=x-y$. Furthermore, increasing the kernel size $\sigma$ makes correntropy tends to the correlation of $x$ and $y$~\cite{santamaria2006generalized}.

A similar quantity to the correntropy is the centered correntropy:
\begin{equation}
\begin{aligned}
U(x,y) & = \mathbb{E}[\kappa(x,y)] - \mathbb{E}_x\mathbb{E}_y[\kappa(x,y)] \\
&= \iint \kappa(x,y)(dF_{X,Y}(x,y)-dF_X(x)dF_Y(y)),
\end{aligned}
\end{equation}
where $F_X(x)$ and $F_Y(y)$ are the marginal distributions of $X$ and $Y$, respectively.

The centered correntropy can be interpreted as a nonlinear counterpart of the covariance in RKHS~\cite{chen2017kernel}. Moreover, the following property serves as the basis of our test statistic that will be introduced later.

\begin{property}
Given $n$ random variables ${x_1,x_2,\cdots,x_n}$ and any set of real numbers ${\alpha_1,\alpha_2,\cdots,\alpha_n}$, for any symmetric positive definite kernel $\kappa(x,y)$, the centered correntropy matrix $C$ defined as $C(i,j)=U(x_i,x_j)$ is always positive semidefinite, i.e., $C\in \mathcal{S}^n_{+}$.
\end{property}

\begin{proof}
By Property $2$ in~\cite{rao2011test}, $U(x,y)$ is a symmetric positive semidefinite function, from which it follows that:
\begin{equation}
\sum_{i=1}^n\sum_{j=1}^n \alpha_i \alpha_j U(x_i,x_j) \geq 0.
\end{equation}
$C$ is obviously symmetric. This concludes our proof.
\end{proof}

In practice, data distributions are unknown and only a finite number of observations $\{x_i,y_i\}_{i=1}^N$ are available, which leads to the sample estimator of centered correntropy\footnote{Throughout this work, we determine kernel width $\sigma$ with the Silverman's rule of thumb~\cite{Silverman86}.}~\cite{rao2011test}:
\begin{equation}
\hat{U}(x,y) = \frac{1}{N}\sum_{i=1}^N\kappa_\sigma(x_i,y_i) - \frac{1}{N^2}\sum_{i}^N\sum_{j}^N\kappa_\sigma(x_i,y_j).
\end{equation}

\section{Methods}
\subsection{Problem Formulation}


We have two groups of observations $\{(x_i^1,y_i^1)\}_{i=1}^{N_1}$ and $\{(x_i^2,y_i^2)\}_{i=1}^{N_2}$ that are assumed to be independently and identically distributed ($i.i.d.$) with density functions $p_1(x,y)$ and $p_2(x,y)$, respectively. $y$ is a dependent variable that takes values in $\mathbb{R}$, and $x$ is a vector of explanatory variables that takes values in $\mathbb{R}^p$.
Typically, the conditional distribution $p(y|x)$ are unknown and unspecified. The aim of this paper is to suggest a test statistic to measure the nearness between $p_1(y|x)$ and $p_2(y|x)$.
\begin{eqnarray*}
H_0: \mathsf{Pr}\left(p_1(y|x)=p_2(y|x)\right)=1\\
H_1: \mathsf{Pr}\left(p_1(y|x)=p_2(y|x)\right)<1
\end{eqnarray*}

\subsection{Our Statistic and the Conditional Test}

We define the divergence from $p_1(y|x)$ to $p_2(y|x)$ as:
\begin{equation}\label{eq:VN_conditional_divergence}
\begin{aligned}
D_{\varphi,B}(p_1(y|x)\|p_2(y|x)) &=& D_{\varphi,B}(C_{xy}^1\|C_{xy}^2) \\ &&  - D_{\varphi,B}(C_{x}^1\|C_{x}^2),
\end{aligned}
\end{equation}
where $C_{xy}\in \mathcal{S}_{+}^{p+1}$ denotes the centered correntropy matrix of the random vector concatenated by $x$ and $y$, and $C_x \in \mathcal{S}_{+}^p$ denotes the centered correntropy matrix of $x$. Obviously, $C_x$ is a submatrix of $C_{xy}$ by removing the row and column associated with $y$.

Although Eq.~(\ref{eq:VN_conditional_divergence}) is assymetic itself, one can easily achieve symmetry by taking the form:
\begin{equation}\label{eq:VN_conditional_symmetric}
\begin{aligned}
D_{\varphi,B}(p_1(y|x):p_2(y|x)) & = \frac{1}{2}\large(D_{\varphi,B}(p_1(y|x)\|p_2(y|x)) \\
& + D_{\varphi,B}(p_2(y|x)\|p_1(y|x))\large).
\end{aligned}
\end{equation}


\begin{algorithm}[htb]
\caption{Test the conditional distribution divergence (CDD) based on the matrix Bregman divergence}
\label{PermutationAlg}
\begin{algorithmic}[1]
\Require
Two groups of observations $S^1 = \{(x_i^1,y_i^1)\}_{i=1}^{N_1}$ and $S^2 = \{(x_i^2,y_i^2)\}_{i=1}^{N_2}$, $x_i\in \mathbb{R}^p$, $y_i\in \mathbb{R}$;
$D_{\varphi,B}$;
Permutation number $P$;
Significant rate $\eta$.
\Ensure
Test \emph{decision} (Is $H_0: \mathsf{Pr}\left(p_1(y|x)=p_2(y|x)\right)=1$ $True$ or $False$?).
\State Measure CDD $d_{0}$ on $S^1$ and $S^2$ with Eq.~(\ref{eq:VN_conditional_symmetric}).
\For {$t = 1$ to $P$}
\State $(S^1_t, S^2_t)\leftarrow$ random split of $S^1\bigcup S^2$.
\State Measure CDD $d_{t}$ on $S^1_t$ and $S^2_t$ with Eq.~(\ref{eq:VN_conditional_symmetric}).
\EndFor
\If {$\frac{1+\sum\nolimits_{t=1}^P\mathbf{1}[d_{0}\leq d_t]}{1+P}\leq\eta$}
\State \emph{decision}$\leftarrow$$False$
\Else
\State \emph{decision}$\leftarrow$$True$
\EndIf\\
\Return \emph{decision}
\end{algorithmic}
\end{algorithm}

Given Eq.~(\ref{eq:VN_conditional_divergence}) and Eq.~(\ref{eq:VN_conditional_symmetric}), we design a simple permutation test to distinguish $p_1(y|x)$ from $p_2(y|x)$. Our test methodology is shown in Algorithm~\ref{PermutationAlg}, where $\mathbf{1}$ indicates an indicator function.
The intuition behind this scheme is that if there is no difference on the underlying conditional distributions, the test statistic on the ordered split (i.e., $d_0$) should not deviate too much from that of the shuffled splits (i.e., $\{d_t\}_{t=1}^P$).

\section{Conditional Divergence Properties}

We present useful properties of our statistic (i.e., Eq.~(\ref{eq:VN_conditional_divergence}) or Eq.~(\ref{eq:VN_conditional_symmetric})) based on the LogDet divergence. In particular we show it is non-negative (when evaluated on covariance matrices) which reduces to zero when two data sets share the same linear regression function. We also perform Monte Carlo simulations to investigate its detection power and establish its connection to prior art.

\begin{property}
    $D_{\ell d}(\Sigma_{xy}^1||\Sigma_{xy}^2) - D_{\ell d}(\Sigma_{x}^1||\Sigma_{x}^2) \ge 0 $.
\end{property}
\begin{proof}
The proof follows immediately from the fact that $D_{\ell d}(\Sigma_{yx}^1||\Sigma_{yx}^2) - D_{\ell d}(\Sigma_{x}^1||\Sigma_{x}^2)$ is the KL divergence between $p_1(y|x)$ and $p_2(y|x)$ for Gaussian distributions with zero mean and of covariances $\Sigma_{yx}^1,\Sigma_{yx}^2$ (see Eq.~(\ref{eq:KL_conditional_decomposition})) and $D_{KL}(p_1(y|x)||p_2(y|x))\ge 0$.
\end{proof}

\begin{property} \label{eq:zero}
	Let $x_1 \sim N(\mu_x^1,\Sigma_x^1)$ and $x_2 \sim N(\mu_x^2,\Sigma_x^2) $ be two input processes of full rank covariance matrices of size $p \times p$. For a common linear system, defined by a full rank matrix $W$ of size $r \times p $ such that $y = W x$, we have:
	$$
	D_{\ell d}(\Sigma_{xy}^1||\Sigma_{xy}^2) - D_{\ell d}(\Sigma_{x}^1||\Sigma_{x}^2) = 0
	$$
\end{property}

\begin{proof}
Denote $M = \left| \begin{array}{c}
I_p \\ W
\end{array}\right|$, where $I_p$ denotes an identity matrix, we have $\Sigma_{xy}^i = W\Sigma_x^iW^T$, from which we have:
\begin{eqnarray*}
D_{\ell d}(\Sigma_{xy}^1||\Sigma_{xy}^2) - D_{\ell d}(\Sigma_{x}^1||\Sigma_{x}^2) &=& \\
D_{\ell d}(M\Sigma_{x}^1M^T||M\Sigma_{x}^2M^T) - D_{\ell d}(\Sigma_{x}^1||\Sigma_{x}^2) &= & 0,
\end{eqnarray*}
where we used Property 12 and Lemma 5 of \cite{kulis2009low}, since $\range (\Sigma_{xy}^i) = p \le r + p$.
\end{proof}

\subsection{Power Test}
Our aim here is to examine if our statistic is really suitable for quantifying the discrepancy between two conditional distributions. Motivated by~\cite{zheng2000consistent,fan2006nonparametric}, we generate four groups of data that have distinct conditional distributions. Specifically, in model (a), the dependent variable $y$ is generated by $y = 1 + \sum_{i=1}^p x_i + \epsilon$, where $\epsilon$ denotes standard normal distribution. In model (b), $y = 1 + \sum_{i=1}^p x_i + \psi$, where $\psi$ denotes the standard Logistic distribution. In model (c), $y = 1 + \sum_{i=1}^p \log x_i + \epsilon$. In model (d), $y = 1 + \sum_{i=1}^p \log x_i + \psi$. For each model, the input distribution is an isotropic Gaussian.

To evaluate the detection power of our statistic on any two models, we randomly generate $500$ samples from each model which has the same dimensionality $m$ on explanatory variable $x$. We then use Algorithm~\ref{PermutationAlg} ($P=500,\eta=0.1$) to test if our statistic can distinguish these two data sets. We repeat this procedure with $100$ independent runs and use the percentage of success as the detection power. The conditional KL divergence (see Eq.~(\ref{eq:KL_conditional_divergence})) estimated with an adaptive $k$NN estimator~\cite{wang2009divergence} is implemented as a baseline competitor.
Table~\ref{tab:power_test} summarizes the power test results when $p=3$ and $p=30$. Although all methods perform good for $p=3$, our test statistic is significantly more powerful than conditional KL divergence in high-dimensional space.

\begin{figure}[ht]
\centering
\begin{subfigure}[]{0.48\textwidth}
\includegraphics[width=\textwidth,height = 3cm]{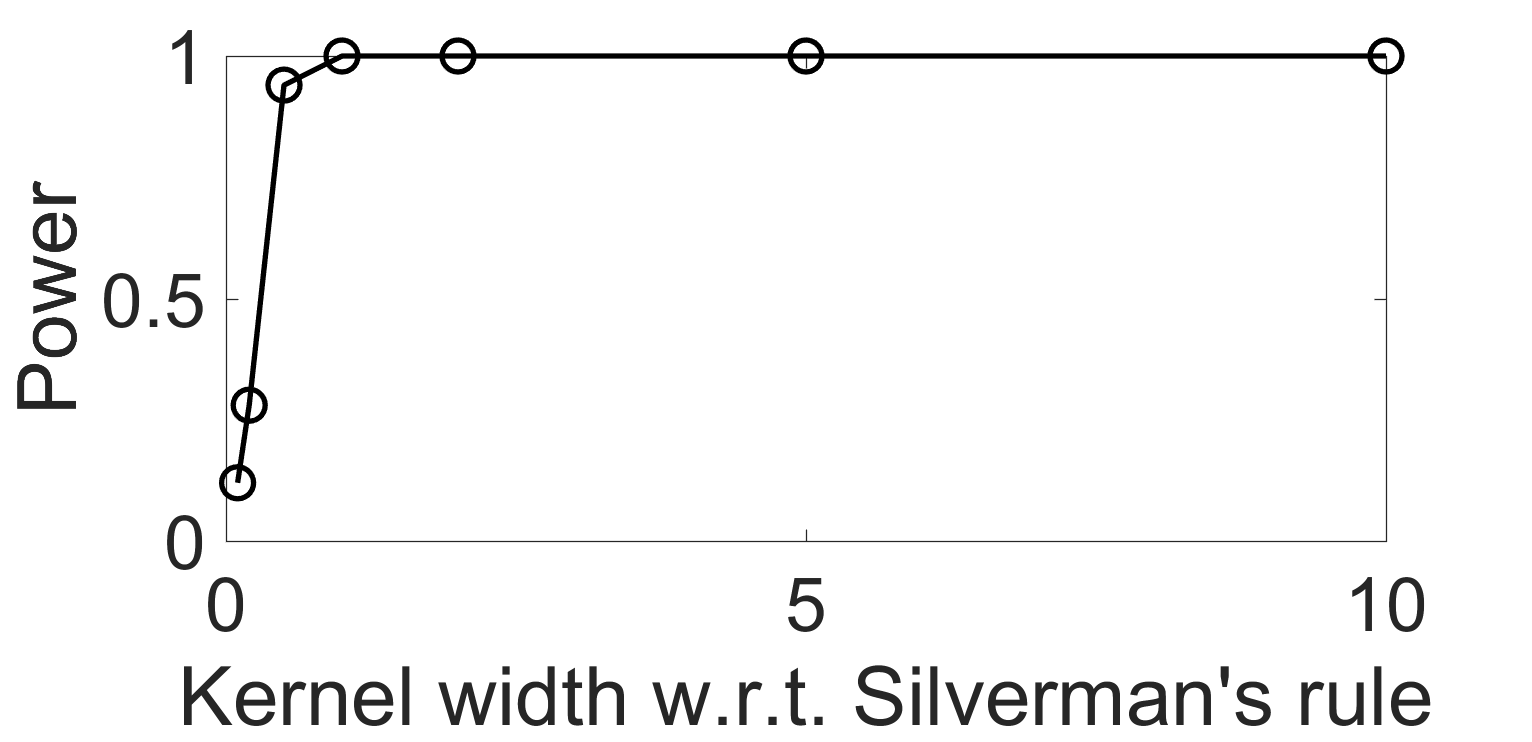}
\end{subfigure} \\
\begin{subfigure}[]{0.48\textwidth}
\includegraphics[width=\textwidth,height = 3cm]{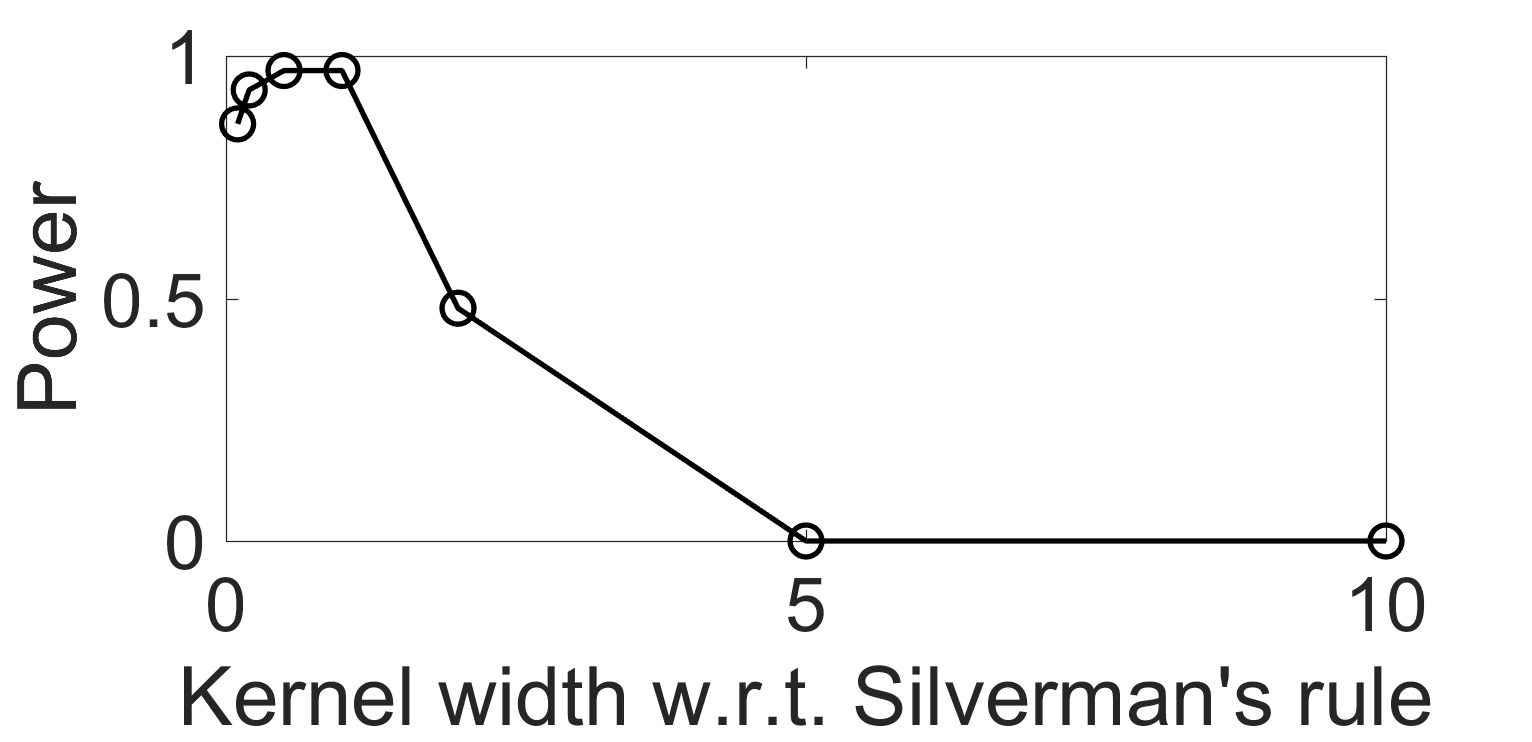}
\end{subfigure}
\caption{Power of our statistics with respect to the kernel width. The $x$-axis denotes the ratio of our used kernel width with respect to the one selected with Silverman's rule of thumb.}
\label{fig:power_test}
\end{figure}

We also depict in Fig.~\ref{fig:power_test} the power of our statistic in case I: model (a) against model (b) and case II: model (c) against model (d), with respect to different kernel widths. For case I (the first row), larger width is preferred. This is because the second-order information dominates in the linear model. For case II (the second row), we need smaller width to capture more higher order information in highly nonlinear model.

\begin{table*}\centering
\ra{1.0}
\begin{tabular}{@{}rrrrrrrrrrrrrrr@{}}\toprule
& \multicolumn{4}{c}{Conditional KL} & \phantom{abc}& \multicolumn{4}{c}{von Neumann ($C$)} &
\phantom{abc} & \multicolumn{4}{c}{LogDet ($C$)}\\
\cmidrule{2-5} \cmidrule{7-10} \cmidrule{12-15}
& (a) & (b) & (c) & (d) && (a) & (b) & (c) & (d) && (a) & (b) & (c) & (d)\\ \midrule
$p=3$\\
(a) & 0.09 & 1 & 1 & 1 && 0.04 & 1 & 1 & 1 && 0.06 & 1 & 1 & 1\\
(b) & 1 & 0.07 & 1 & 1 && 1 & 0.07 & 1 & 0.96 && 1 & 0.09 & 1 & 0.92\\
(c) & 1 & 1 & 0.12 & 1 && 1 & 1 & 0.13 & 0.97 && 1 & 1 & 0.12 & 0.91\\
(d) & 1 & 1 & 1 & 0.07 && 1 & 0.96 & 0.97 & 0.09 && 1 & 0.94 & 0.97 & 0.07\\
$p=30$\\
(a) & 0.12 & 0.67 & 1 & 1 && 0.04 & 1 & 1 & 1 && 0.10 & 1 & 1 & 1\\
(b) & 0.60 & 0.14 & 1 & 1 && 1 & 0.10 & 1 & 1 && 1 & 0.10 & 1 & 1\\
(c) & 1 & 1 & 0.09 & 0.34 && 1 & 1 & 0.11 & 0.67 && 1 & 1 & 0.06 & 0.60\\
(d) & 1 & 1 & 0.34 & 0.07 && 1 & 1 & 0.58 & 0.14 && 1 & 1 & 0.50 & 0.13\\
\bottomrule
\end{tabular}
\caption{Power test for conditional KL divergence and our statistics implemented with von Neumann and LogDet divergences on centered correntropy matrix $C$}
\label{tab:power_test}
\end{table*}

\subsection{Relation to Previous Efforts}

\subsubsection{Gaussian Data}
As mentioned earlier, the centered correntropy matrix $C$ evaluated with a Gaussian kernel encloses all even higher order statistics of pairwise dimensions of data. If we replace $C$ with its second-order counterpart (i.e., the covariance matrix $\Sigma$), we have:
\begin{equation}\label{eq:VN_conditional_baseline}
D_{\varphi,B}(p_1(y|x)\|p_2(y|x)) = D_{\varphi,B}(\Sigma_{xy}^1\|\Sigma_{xy}^2) - D_{\varphi,B}(\Sigma_{x}^1\|\Sigma_{x}^2).
\end{equation}

Taking Eq.~(\ref{eq:LogDet}) into Eq.~(\ref{eq:VN_conditional_baseline}), we obtain:
\begin{equation}\label{eq:LogDet_conditional_decomposition}
\begin{aligned}
D_{\ell D}(p_1(y|x)\|p_2(y|x))& = \Tr({(\Sigma_{xy}^2)}^{-1}\Sigma_{xy}^1) + \log\frac{|\Sigma_{xy}^2|}{|\Sigma_{xy}^1|} \\ - (p+1)
& - \Tr({(\Sigma_{x}^2)}^{-1}\Sigma_{x}^1) - \log\frac{|\Sigma_{x}^2|}{|\Sigma_{x}^1|} + p.
\end{aligned}
\end{equation}

On the other hand, the KL divergence between two conditional distributions on a pair of random variables satisfies the following chain rule (see proof in Chapter 2 of~\cite{cover2012elements}):
\begin{eqnarray}\label{eq:KL_conditional_divergence}
D_{KL}(p_1(y|x)\|p_2(y|x)) = D_{KL}(p_1(x,y)\|p_2(x,y)) \nonumber \\
- D_{KL}(p_1(x)\|p_2(x)).
\end{eqnarray}

Suppose the data is Gaussian distributed, i.e., $p_1(x)\sim\mathcal{N}(\mu_x^1,\Sigma_x^1)$, $p_2(x)\sim\mathcal{N}(\mu_x^2,\Sigma_x^2)$, $p_1(x,y)\sim\mathcal{N}(\mu_{xy}^1,\Sigma_{xy}^1)$,
$p_2(x,y)\sim\mathcal{N}(\mu_{xy}^2,\Sigma_{xy}^2)$,
Eq.~(\ref{eq:KL_conditional_divergence}) has a closed-form expression (see proof in~\cite{duchi2007derivations}):
\begin{equation}\label{eq:KL_conditional_decomposition}
\begin{aligned}
& D_{KL}(p_1(y|x)\|p_2(y|x)) \\
&= \frac{1}{2}\{ \Tr({(\Sigma_{xy}^2)}^{-1}\Sigma_{xy}^1) + \log_2\frac{|\Sigma_{xy}^2|}{|\Sigma_{xy}^1|} - (p+1) \\
& - \Tr({(\Sigma_{x}^2)}^{-1}\Sigma_{x}^1) - \log\frac{|\Sigma_{x}^2|}{|\Sigma_{x}^1|} + p \\
& +
(\mu_{xy}^2 - \mu_{xy}^1)^T(\Sigma_{xy}^2)^{-1}(\mu_{xy}^2 - \mu_{xy}^1)  \\
&  - (\mu_{x}^2 - \mu_{x}^1)^T(\Sigma_{x}^2)^{-1}(\mu_{x}^2 - \mu_{x}^1)\}.
\end{aligned}
\end{equation}

Comparing Eq.~(\ref{eq:LogDet_conditional_decomposition}) with Eq.~(\ref{eq:KL_conditional_decomposition}), it is easy to find that our baseline variant reduces to the conditional KL divergence under Gaussian assumption. The only difference is that the conditional KL divergence contains a Mahalanobis Distance term on the mean, which can be interpreted as the first-order information of data.

\subsubsection{Beyond Gaussian Data}
Gaussian assumption is always over-optimistic. If we stick to the conditional KL divergence, the probability estimation becomes inevitable, which is notoriously difficult in high-dimensional space~\cite{nagler2016evading}.
By making use the correntropy function, our statistic avoids the estimation of the underlying distribution, but it explicitly incorporates the higher order information which was lost in Eq.~(\ref{eq:VN_conditional_baseline}).

\section{Machine Learning Applications}
We present three solid examples on machine learning applications to demonstrate the performance improvement in the state-of-the-art (SOTA) methodologies gained by our conditional divergence statistic. 

\subsection{Multitask Learning}

Consider an input set $\mathcal{X}$ and an output set $\mathcal{Y}$ and for simplicity that $\mathcal{X}\in \mathbb{R}^p$, $\mathcal{Y}\in \mathbb{R}$. Tasks can be viewed as $T$ functions $f_t,t=1,\cdots,T$, to be learned from given data $\{(x_{ti},y_{ti}):i=1,\cdots,N_t, t=1,\cdots,T\}\subseteq \mathcal{X}\times \mathcal{Y}$, where $N_t$ is the number of samples in the $t$-th task. These tasks may be viewed as drawn from an unknown joint distribution of tasks, which is the source of the bias that relates the tasks. Multitask learning is the paradigm that aims at improving the generalization performance of multiple prediction problems (tasks) by exploiting potential useful information between related tasks.

\subsubsection{Visualizing Task-relatedness}
We first test if our proposed statistic is able to reveal the relatedness among multiple tasks. To this end, we select data from $29$ tasks that are collected from various landmine fields\footnote{\url{http://www.ee.duke.edu/~lcarin/LandmineData.zip}.}. Each object in a given data set is represented by a $9$-dimensional feature vector and the corresponding binary label ($1$ for landmine and $0$ for clutter). The landmine detection problem is thus modeled as a binary classification problem.



\begin{figure}[ht]
  \begin{subfigure}[]{0.23\textwidth}
    \includegraphics[width=\textwidth]{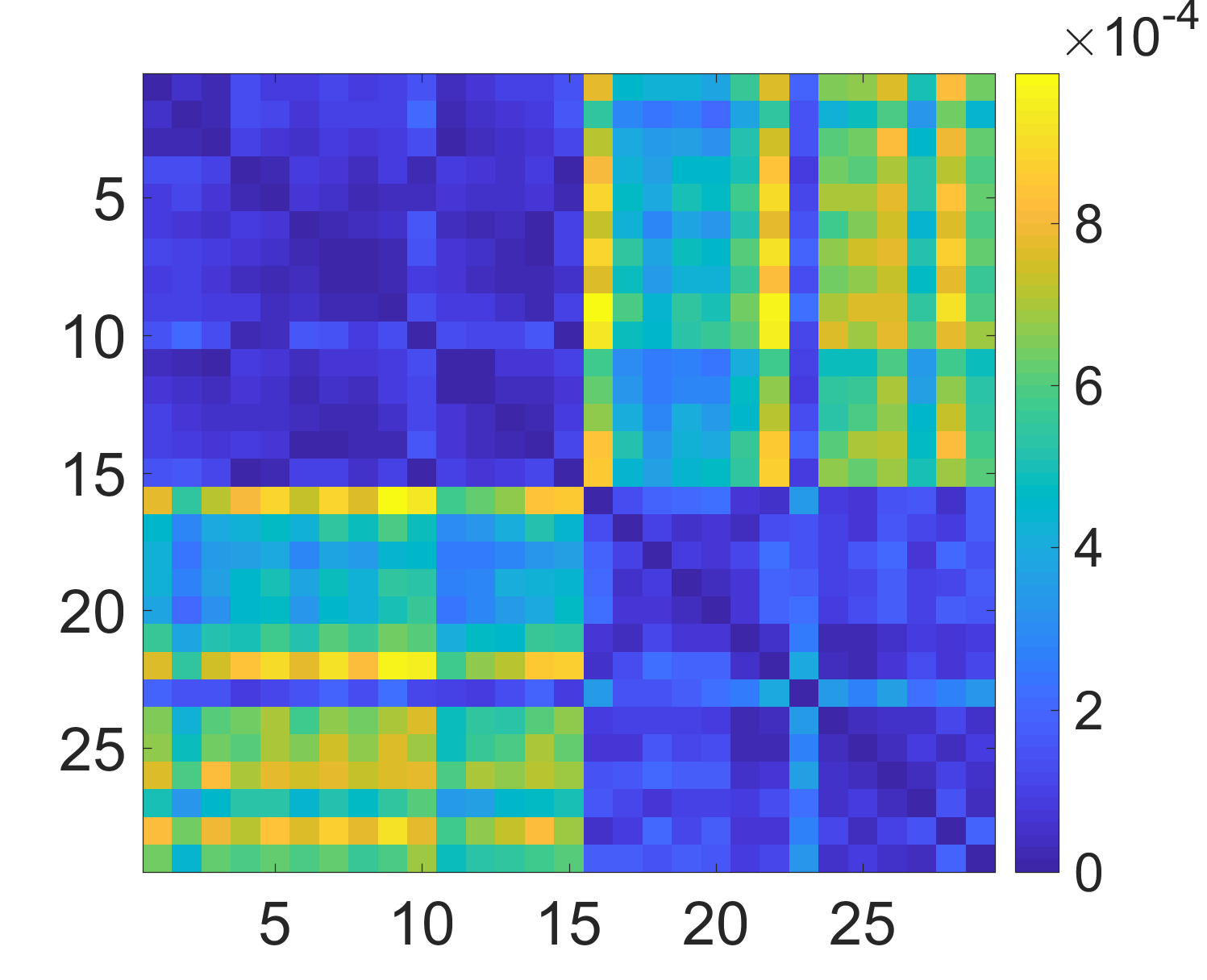}
    \caption{vN ($C$)}
  \end{subfigure}
  \begin{subfigure}[]{0.23\textwidth}
    \includegraphics[width=\textwidth]{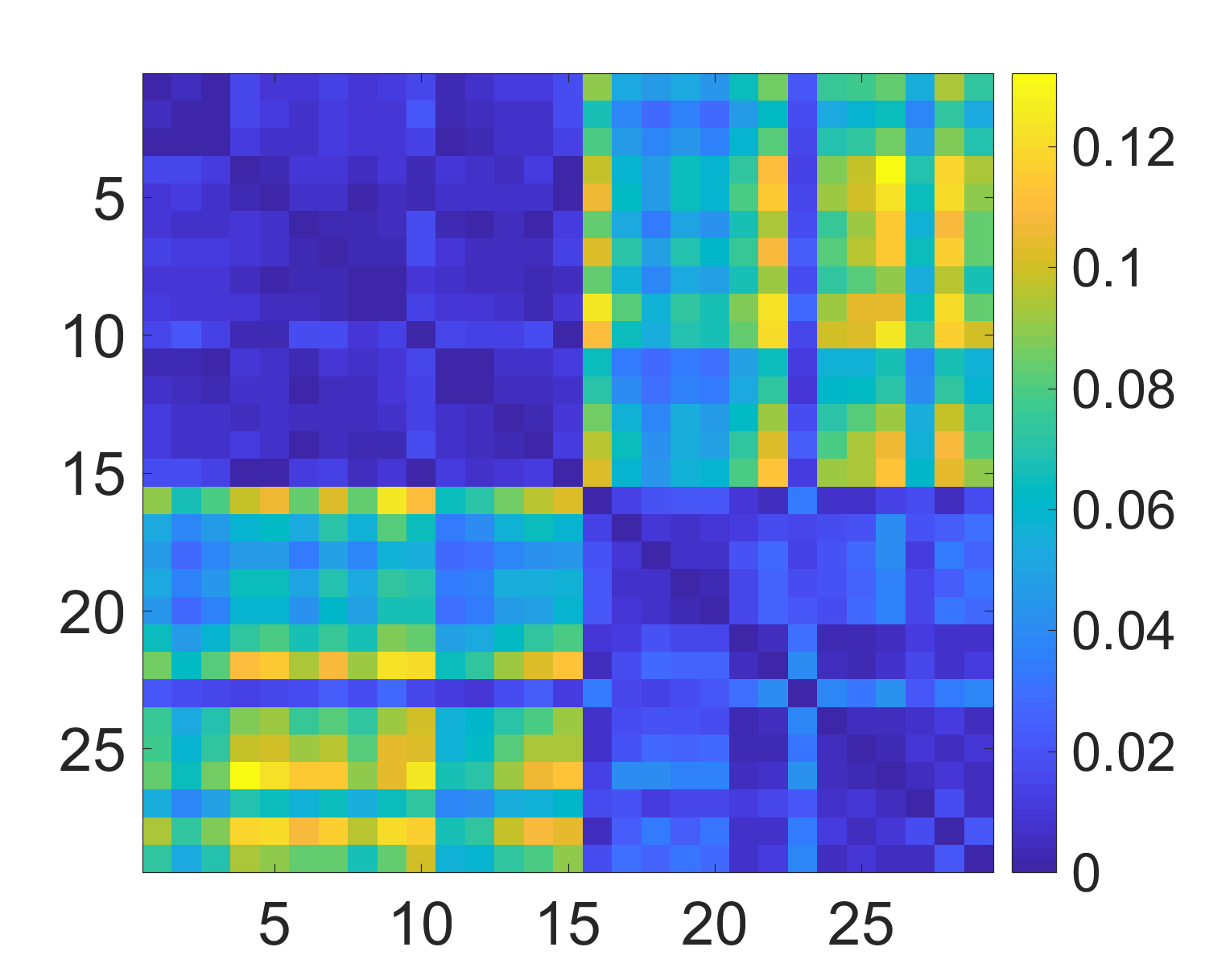}
    \caption{vN ($\Sigma$)}
  \end{subfigure} \\
    \begin{subfigure}[]{0.23\textwidth}
    \includegraphics[width=\textwidth]{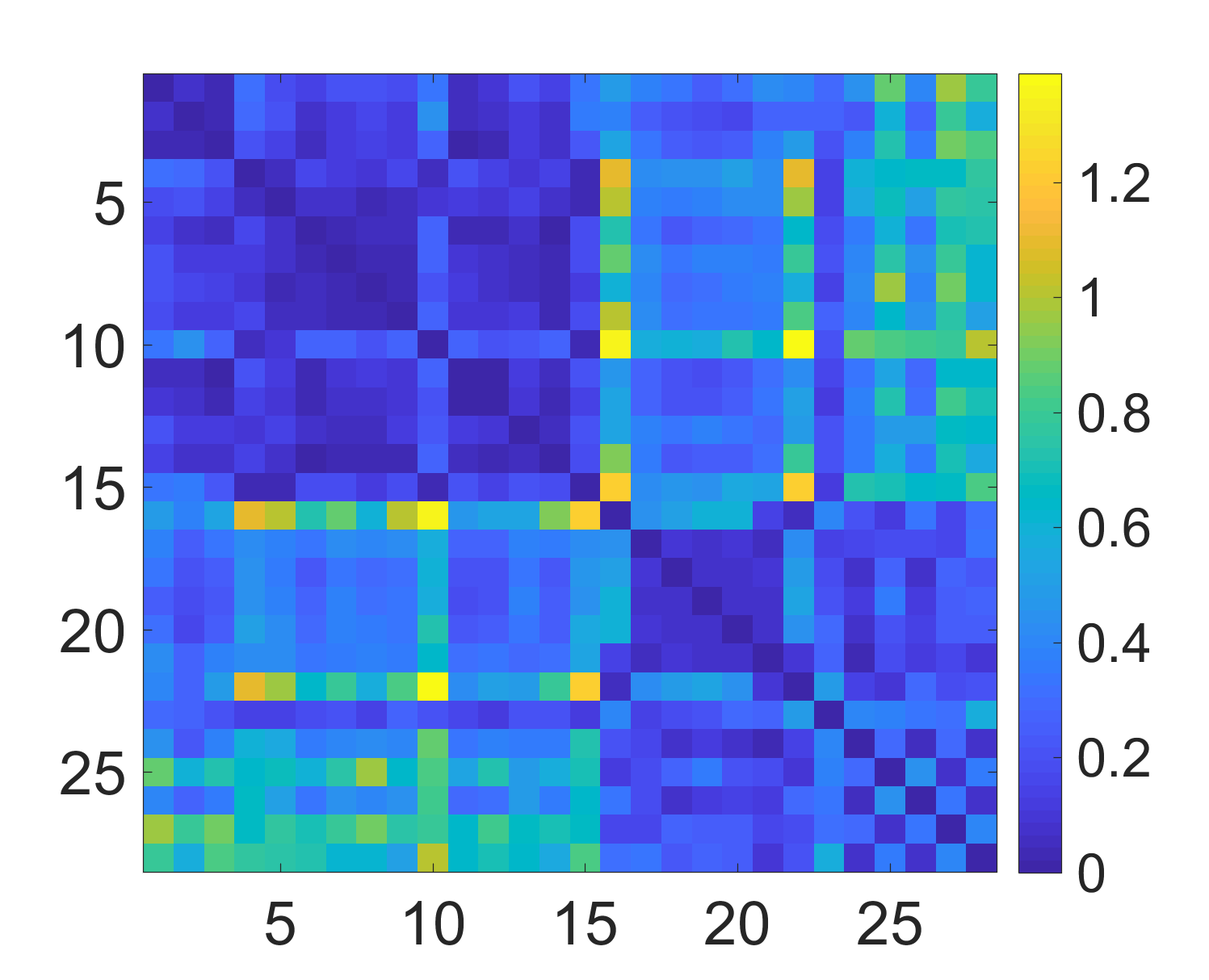}
    \caption{LogDet ($C$)}
  \end{subfigure}
    \begin{subfigure}[]{0.23\textwidth}
    \includegraphics[width=\textwidth]{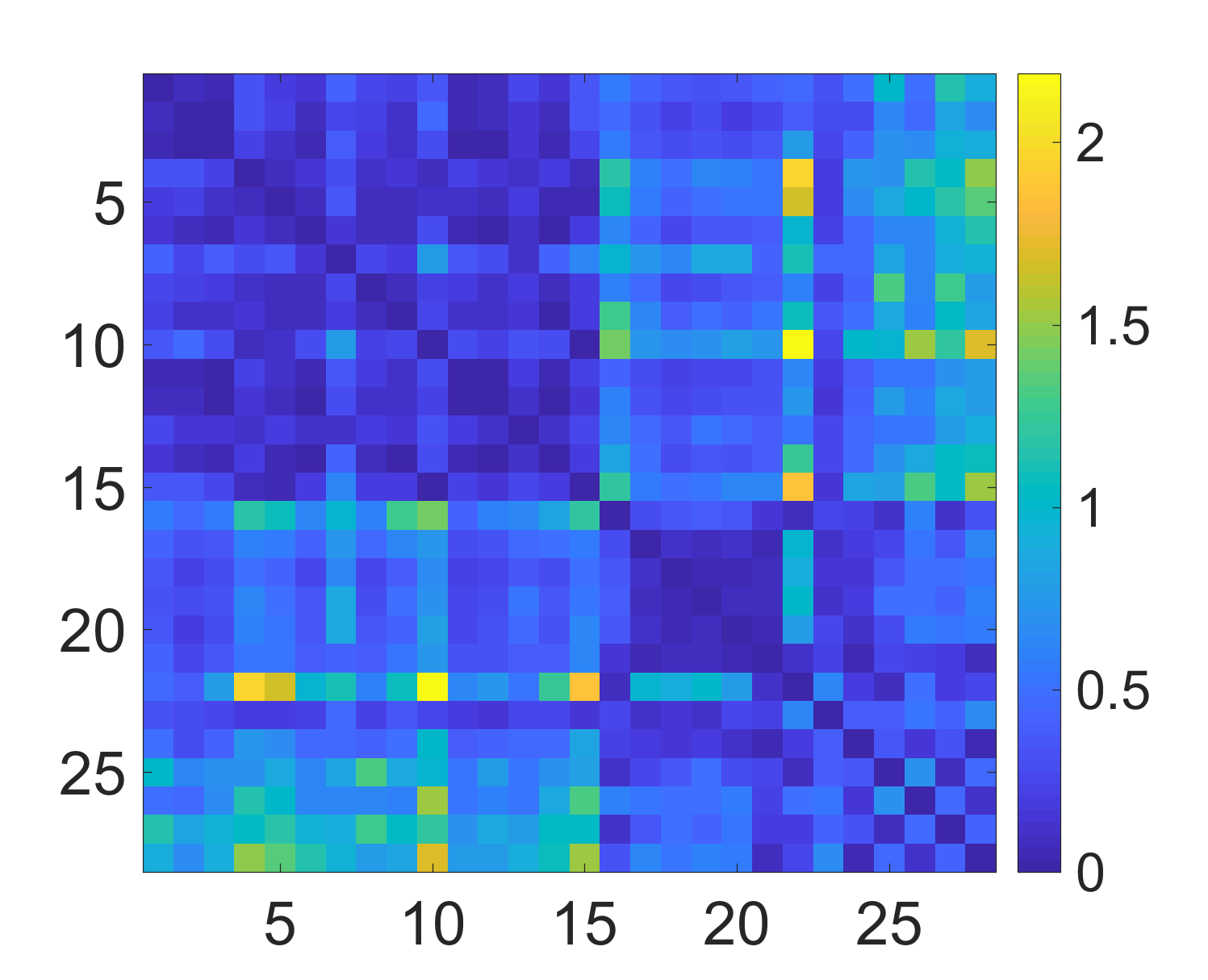}
    \caption{LogDet ($\Sigma$)}
  \end{subfigure} \\
    \begin{subfigure}[]{0.23\textwidth}
    \includegraphics[width=\textwidth]{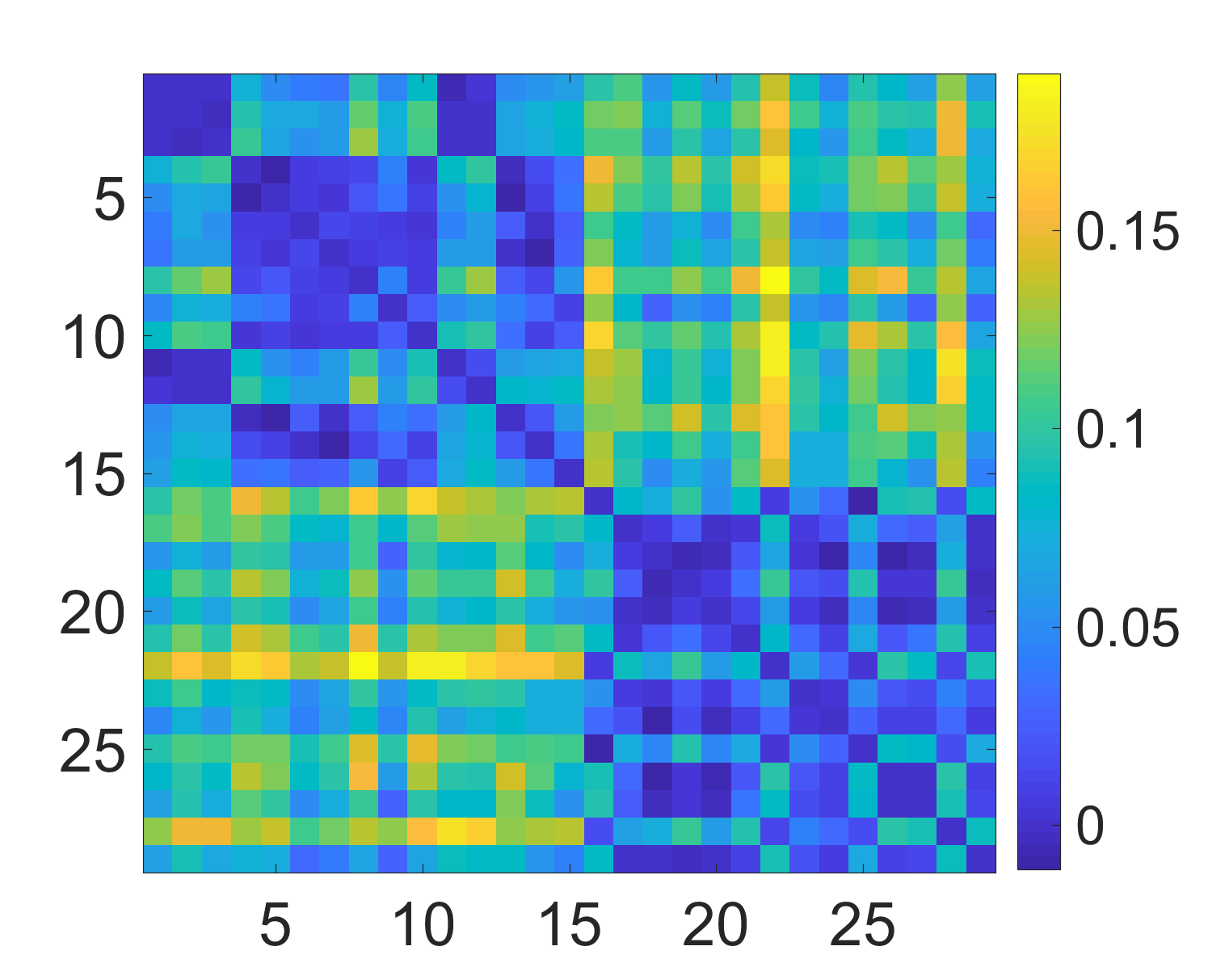}
    \caption{Conditional KL}
  \end{subfigure}
  \begin{subfigure}[]{0.23\textwidth}
    \includegraphics[width=\textwidth]{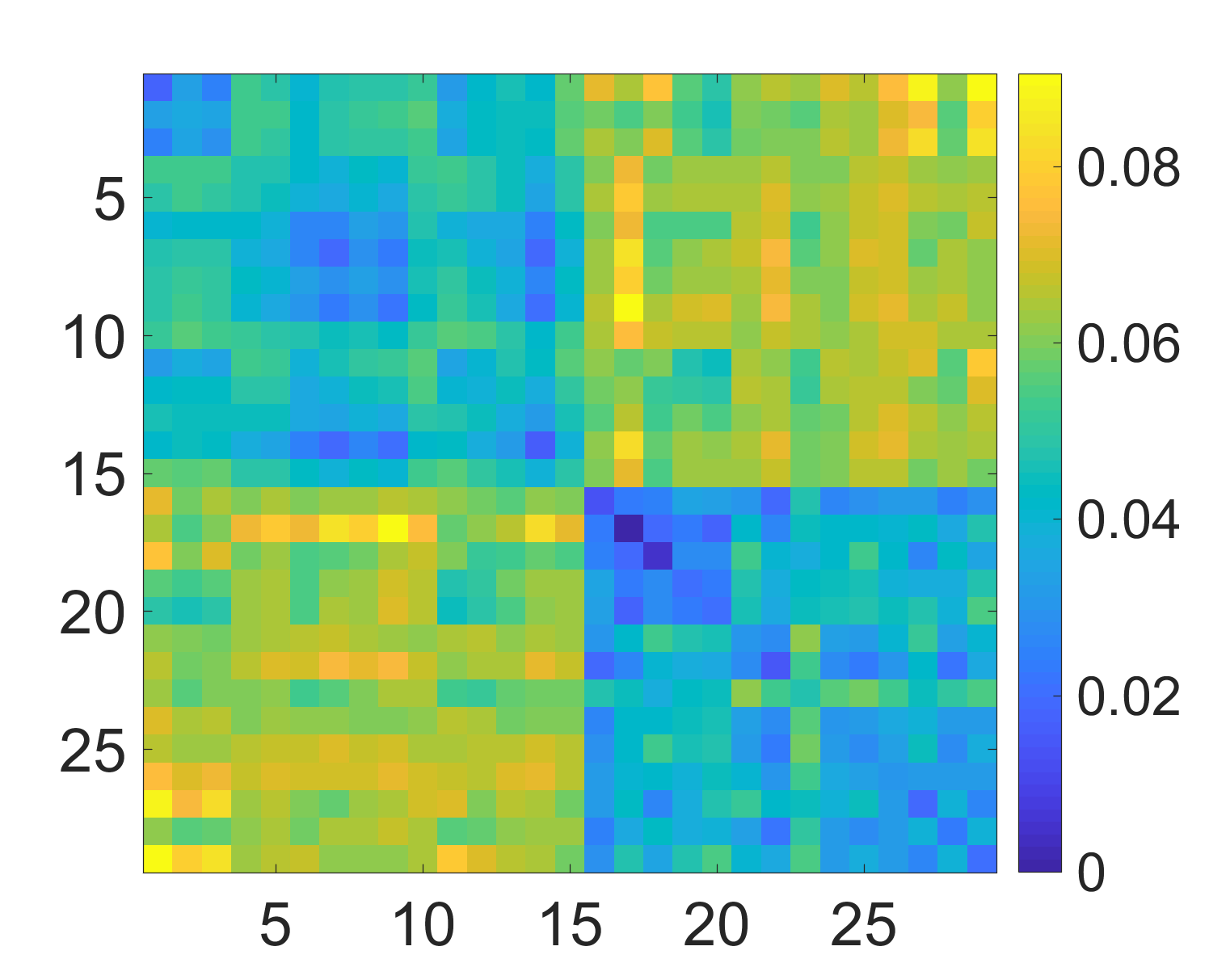}
    \caption{MTRL}
  \end{subfigure}
  \caption{Visualize task-relatedness in landmine detection data set. ``vN" refers to von Neumann, $C$ denotes centered correntropy matrix, and $\Sigma$ denotes covariance matrix. For example, vN (C) denotes our statistic implemented with von Neumann divergence on centered correntropy matrix.}
  \label{fig:div_com}
\end{figure}

Among these $29$ data sets, $1$-$15$ correspond to regions that are relatively highly foliated and $16$-$29$ correspond to regions that are bare earth or desert. We measure the task-relatedness with the conditional divergence $D_{\varphi,B}(p_1(y|x):p_2(y|x))$ and demonstrate their pairwise relationships in a task-relatedness matrix.
Thus we expect that there are approximately two clusters in the task-relatedness matrix corresponding to two classes of ground surface condition. Fig.~\ref{fig:div_com} demonstrates the visualization results generated by our proposed statistic and the conditional KL divergence estimated with an adaptive $k$NN estimator~\cite{wang2009divergence}. We also compare our method with MTRL~\cite{zhang2010convex}, a widely used objective to learn multiple tasks and their relationships in a convex manner. Our vN (C) and vN ($\Sigma$) clearly demonstrate two clusters of tasks. By contrast, both MTRL and conditional KL divergence indicate a few misleading relations (i.e., tasks in the same group surface condition have high divergence values).

\subsubsection{Multitask Learning over Graph Structure}
In the second example, we demonstrate how our statistic improves the performance of the Convex Clustering Multi-Task Learning (CCMTL)~\cite{he2019efficient}, a SOTA method for learning multiple regression tasks. The learning objective of CCMTL is given by:
\begin{equation}
\min_{W}\frac{1}{2}\sum_{t=1}^T||w_t^Tx_t-y_t||_2^2 + \frac{\lambda}{2}\sum_{i,j \in G}||w_i - w_j||_2 ,
\label{eq:obj}
\end{equation}
where $W=[w_1^T,w_2^T,\cdots,w_T^T]\in \mathbb{R}^{T\times p}$ is the weight matrix constitutes of the learned linear regression coefficients in each task, $G$ is the graph of relations over all tasks (if two tasks are related, then there is an edge to connect them), and $\lambda$ is a regularization parameter.

CCMTL requires an initialization of the graph structure $G_0$. In the original paper, the authors set $G_0$ as a $k$NN graph on the prediction models learned independently for each task. In this sense, the task-relatedness between two tasks is modeled as the $\ell_2$ distance of their independently learned linear regression coefficients.

\begin{figure}[ht]
\centering
\includegraphics[width=0.45\textwidth]{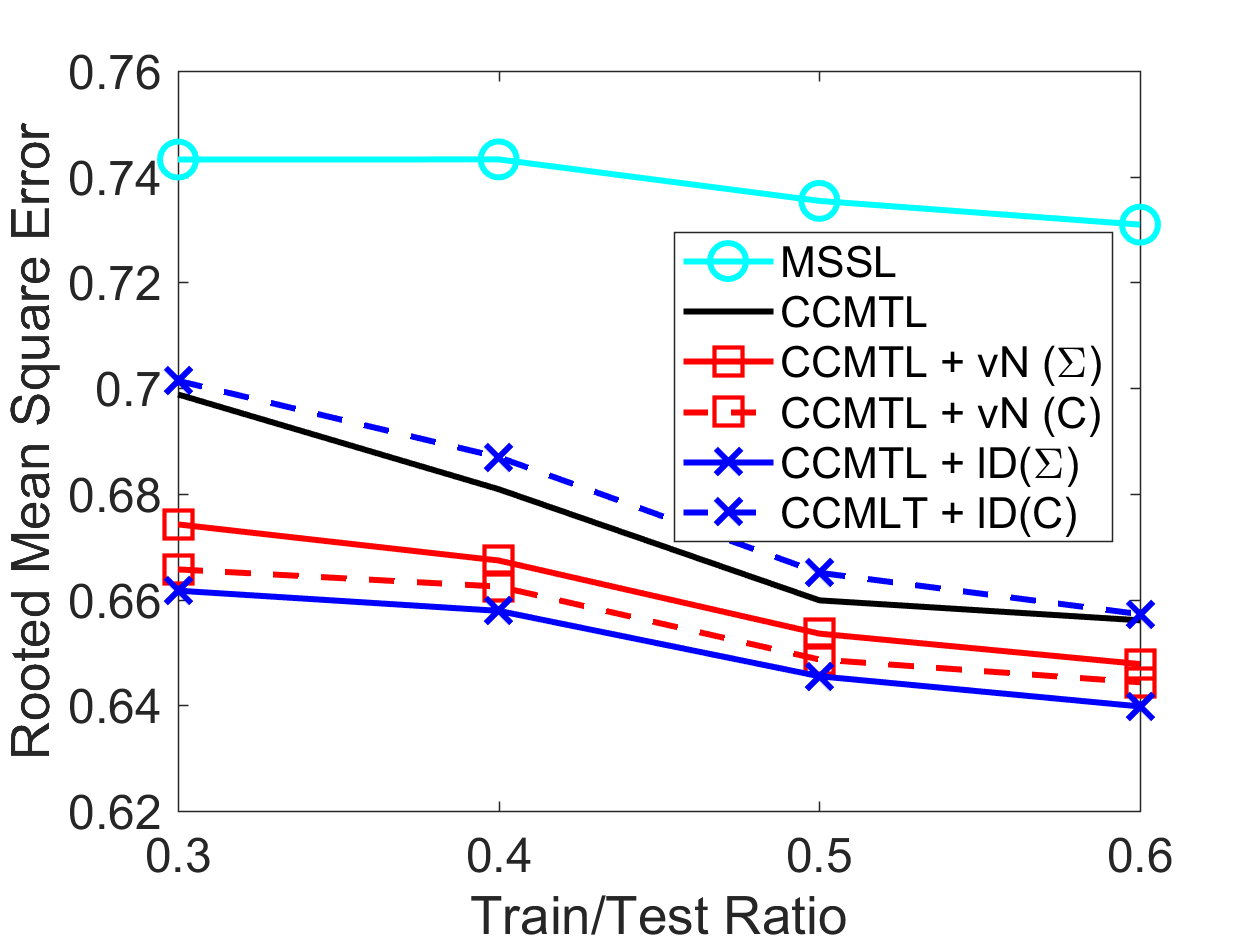}
\caption{Root mean square errors (RMSE) of different multitask learning methodologies with respect to varying train/test ratios on the Parkinson's data set.}
\label{fig:Parkinsons}
\end{figure}

We replace the na\"ive $\ell_2$ distance with our proposed statistic to reconstruct the initial $k$NN graph for CCMTL and test its performance on a real-world Parkinson's disease data set~\cite{tsanas2009accurate}. We have $42$ tasks and $5,875$ observations, where each task and observation correspond to a prediction of the symptom score (motor UPDRS) for a patient and a record of a patient, respectively.
Fig.~\ref{fig:Parkinsons} depicts the prediction root mean square errors (RMSE) under different train/test ratios. To highlight the superiority of CCMTL and our improvement, we also compare it with MSSL~\cite{gonccalves2016multi}, a recently proposed alternative objective to MTRL that learns multiple tasks and their relationships with a Gaussian graphical model. The performance of MTRL is relatively poor, and thus omitted here. Obviously, our statistic improves the performance of CCMTL with a large margin, especially when training samples are limited. Note that, we did not observe performance difference between centered correntropy matrix $C$ and covariance matrix $\Sigma$. This is probably because the higher order information is weak or because the Silverman's rule of thumb is not optimal to tune kernel width here.




\subsection{Concept Drift Detection}
One important assumption underlying common classification models is the stationarity of the training data. However, in real-world data stream applications, the joint distribution $p(x,y)$ between the predictor $x$ and response variable $y$ is not stationary but drifting over time. Concept drift detection approaches aim to detect such drifts and adapt the model so as to mitigate the deteriorating classification performance over time.

Formally, the concept drift between time instants $t_0$ and $t_1$ can be defined as $p_{t_1}(x,y)\neq p_{t_2}(x,y)$~\cite{gama2014survey}. From a Bayesian perspective, concept drifts can manifest two fundamental forms of changes: 1) a change in the marginal probability $p_t(x)$ or $p_t(y)$; and 2) a change in the posterior probability $p_t(y|x)$. Although any two-sample test (e.g.,~\cite{gretton2012kernel}) on $p_t(x)$ or $p_t(y)$ is an option to detect concept drift, existing studies tend to prioritize detecting posterior distribution change, because it clearly indicates the optimal decision rule~\cite{dries2009adaptive}.

Error-based methods constitute a main category of existing concept drift detectors. These methods keep track of the online classification error or error-related metrics of a baseline classifier. A significant increase or decrease of the monitored statistics may suggests a possible concept drift. Unfortunately, the performance of existing manually designed error-related statistics depends on the baseline classifier, which makes them either perform poorly across different data distributions or difficult to be extended to the multi-class classification scenario.

Different from prevailing error-based methods, our statistic operates directly on the conditional divergence $D_{\varphi,B}(p_{t_1}(y|x):p_{t_2}(y|x))$, which makes it possible to fundamentally solve the problem of concept drift detection (without any classifier). To this end, we test the conditional distribution divergence in two consecutive sliding windows (using Algorithm~\ref{PermutationAlg}) at each time instant $t$. A reject of the null hypothesis indicates the existence of a concept drift. Note that the same permutation test procedure has been theoretically and practically investigated in PERM~\cite{harel2014concept}, in which the authors use classification error as the test statistic. Interested readers can refer to~\cite{harel2014concept,yu2017concept} for more details on concept drift detection with permutation test.

\begin{table}[h]
    \begin{subtable}[h]{0.45\textwidth}
        \centering
        \begin{tabular}{l | l | l | l | l}
        Method & Precision & Recall & Delay & Accuracy (\%) \\
        \hline \hline
        DDM & 0.49 & 0.50 & 50 & 89.22 \\
        EDDM & 0.69 & 0.82 & 230 & 92.60 \\
        HDDM & $\mathbf{1}$ & 0.83 & 133 & 97.47 \\
        PERM & 0.81 & 0.88 & 99 & $\mathbf{97.81}$ \\ \hline
        vN ($\Sigma$) & 0.77 & $\mathbf{1}$ & $\mathbf{43}$ & 92.82 \\
        LD ($\Sigma$) & 0.83 & $\mathbf{1}$ & 113 & 93.43 \\
        vN ($C$) & 0.80 & $\mathbf{1}$ & 60 & 90.07 \\
        LD ($C$) & 0.77 & $\mathbf{1}$ & 53 & 92.23
       \end{tabular}
       \caption{Digits08}
       \label{tab:week1}
    \end{subtable}
    \hfill
    \begin{subtable}[h]{0.45\textwidth}
        \centering
        \begin{tabular}{l | l | l | l | l}
        Method & Precision & Recall & Delay & Accuracy (\%) \\
        \hline \hline
        DDM & 0.83 & 0.83 & 25 & 67.47 \\
        EDDM & 0.47 & $\mathbf{1}$ & 46 & 63.23 \\
        HDDM & $\mathbf{1}$ & $\mathbf{1}$ & 15 & 76.94 \\
        PERM & 0.50 & $\mathbf{1}$ & 31 & 71.98 \\ \hline
        vN ($\Sigma$) & 0.50 & $\mathbf{1}$ & $\mathbf{11}$ & 72.11 \\
        LD ($\Sigma$) & 0.47 & $\mathbf{1}$ & 21 & 75.94 \\
        vN ($C$) & 0.50 & $\mathbf{1}$ & $\mathbf{11}$ & 72.52  \\
        LD ($C$) & 0.49 & $\mathbf{1}$ & 23 & $\mathbf{77.02}$
        \end{tabular}
        \caption{Abrupt Insect}
        \label{tab:week2}
     \end{subtable}
     \caption{Quantitative metrics on real-world data sets. The Precision, Recall and Delay denote the concept drift detection precision value, recall value and detection delay, whereas the Accuracy denotes the classification accuracy in the testing set (\%).}
     \vspace{-0.7em}
     \label{tab:concept}
\end{table}

We evaluate the performance of our method against four SOTA error-based concept drift detectors (i.e., DDM~\cite{gama2004learning}, EDDM~\cite{baena2006early}, HDDM~\cite{frias2014online}, and PERM) on two real-world data streams, namely the Digits08~\cite{sethi2017reliable} and the AbruptInsects~\cite{dos2016fast}. Among the selected competitors, HDDM represents one of the best-performing detectors, whereas PERM is the most similar one to ours. The concept drift detection results and the stream classification results over $30$ independent runs are summarized in Table~\ref{tab:concept}. Our method always enjoys the highest recall values and the shortest detection delay. Note that, the classification accuracy is not as high as we expected. One possible reason is that the short detection delay makes us do not have sufficient number of samples to retrain the classifier.

\subsection{Feature Selection}\label{sec:feature}
Our final application is information-theoretic feature selection. Given a set of variables $S = \{X_1,X_2,\cdots,X_n\}$, feature selection refers to seeking a small subset of informative variables $S^\star\subset S$, such that $S^\star$ contains the most relevant yet least redundant information about a desired variable $Y$. From an information-theoretic perspective, this amounts to maximize the mutual information term $\mathbf{I}(y;S^\star)$. Suppose we are now given a set of ``useless" features $\tilde{S}$ that has the same size as $S^\star$ but has no predictive power to $y$, Eq.~(\ref{eq:objective_equivalence}) suggests that instead of maximizing $\mathbf{I}(y;S^\star)$, one can resort to maximize $D_{KL}(P(y|S^\star)\|P(y|\tilde{S}))$ as an alternative.

\begin{equation}\label{eq:objective_equivalence}
\begin{aligned}
\mathbf{I}(y;S^\star) & = \iint P(y,S^\star) \log{\frac{P(y,S^\star)}{P(y)P(S^\star)}} \\
& = \iint \bigg(P(y|S^\star)\log{\frac{P(y|S^\star)}{P(y)}}\bigg)P(S^\star) \\
& = \mathbb{E}_{S}[D_{KL}(P(y|S^\star)||P(y))] \\
& = \mathbb{E}_{S}[D_{KL}(P(y|S^\star)||P(y|\tilde{S}))],
\end{aligned}%
\end{equation}
the last line is by our assumption that $\tilde{S}$ has no predictive power to $y$ such that $P(y|\tilde{S})=P(y)$.

Motivated by the generic equivalence theorem between KL divergence and the Bregman divergence~\cite{banerjee2005clustering}, we optimize our statistic in a greedy forward search manner (i.e., adding the best feature at each round) and generate ``useless" feature set $\tilde{S}$ by randomly permutating $S^\star$ as conducted in~\cite{franccois2007resampling}.

\begin{figure}[ht]
  \begin{subfigure}[]{0.42\textwidth}
    \includegraphics[width=\textwidth]{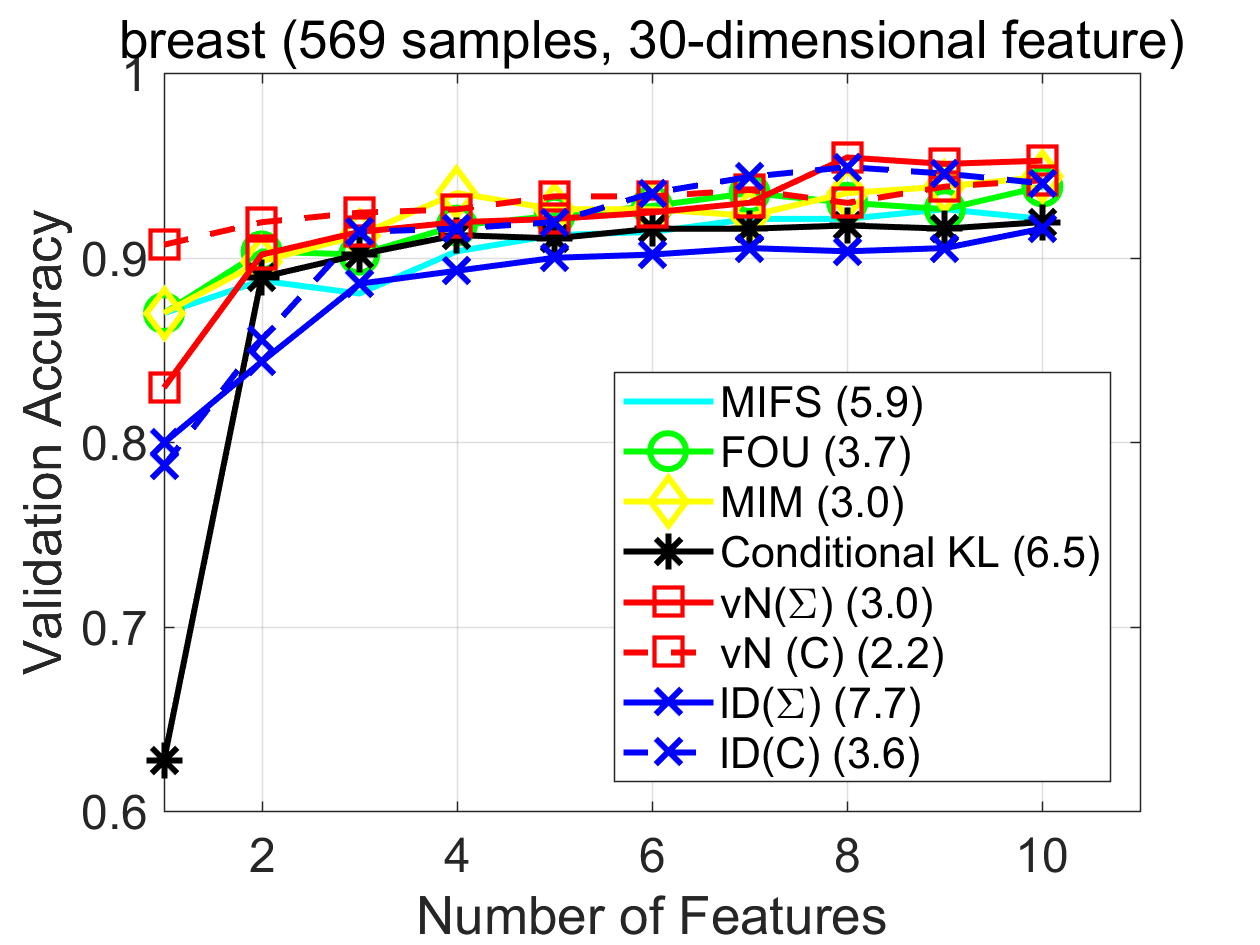}
    \caption{breast}
  \end{subfigure} \\
  \begin{subfigure}[]{0.42\textwidth}
    \includegraphics[width=\textwidth]{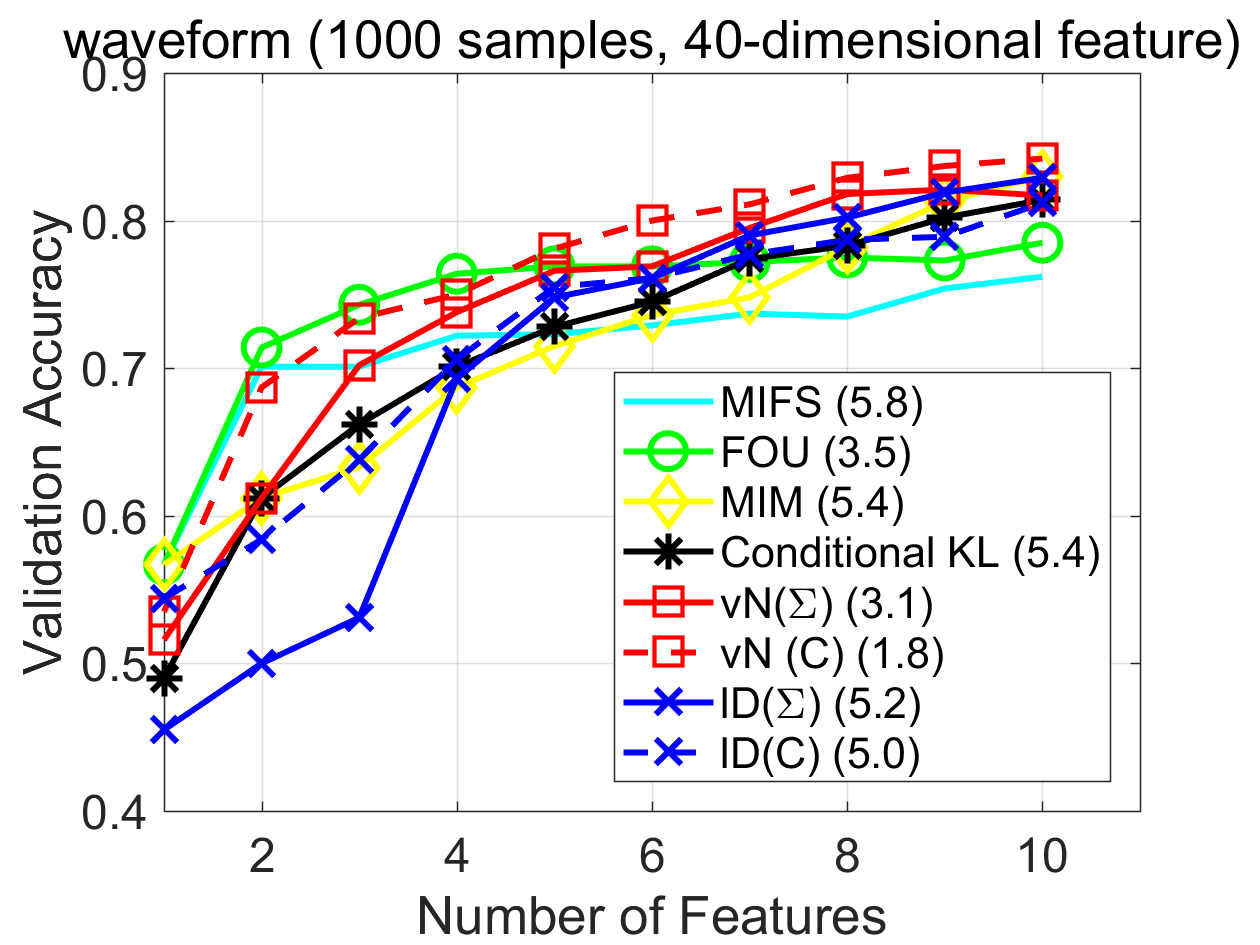}
    \caption{waveform}
  \end{subfigure}
  \caption{Validation accuracy on (a) breast and (b) waveform data sets. The number of samples and the feature dimensionality for each data set are listed in the title. The value beside each method in the legend indicates the average rank in that data set.}
  \label{fig:feature_selection}
\end{figure}

We perform feature selection on two benchmark data sets~\cite{brown2012conditional}. For both data sets, we select $10$ features and use the linear Support Vector Machine (SVM) as the baseline classifier.
We compare our method with three popular information-theoretic feature selection methods that target maximizing $\mathbf{I}(y;S^\star)$, namely MIFS~\cite{battiti1994using}, FOU~\cite{brown2012conditional}, and MIM~\cite{lewis1992feature}. The classification accuracies with respect to different number of selected features (averaged over $10$ fold cross-validation) are presented in Fig.~\ref{fig:feature_selection}. As can be seen, methods on maximizing conditional divergence always achieve comparable performances to those mutual information based competitors. This result confirms Eq.~(\ref{eq:objective_equivalence}). If we look deeper, it is easy to see that our statistic implemented with von Neumann divergence achieves the best performance on both data sets. Moreover, by incorporating higher order information,  centered correntropy performs better than its covariance based counterpart.

\section{Conclusions}
We propose a simple statistic to quantify conditional divergence. We also establish its connections to prior art and illustrate some of its fundamental properties. A natural extension to incorporate higher order information of data is also presented. Three solid examples suggest that our statistic can offer a remarkable performance gain to SOTA learning algorithms (e.g., CCMTL and PERM). Moreover, our statistic enables the development of alternative solutions to classical machine learning problems (e.g., classifier-free concept drift detection and feature selection by maximizing conditional divergence) in a fundamental manner.



Future work is twofold. We will investigate more fundamental properties of our statistic. We will also apply our statistic to more challenging problems, such as the continual learning~\cite{kirkpatrick2017overcoming} which also requires the knowledge of task-relatedness.


\bibliographystyle{named}
\bibliography{von_Neumann_conditional_divergence}

\appendix
\section{Additional Notes (updated in Dec. $27$th $2020$)}

We add a few additional notes to our proposed Bregman-Correntropy (conditional) divergence to illustrate more of its properties in practical applications.

\subsection{The Bregman-Correntropy Divergence $D_{\varphi,B}$ on $p_1(x)$ and $p_2(x)$}

We want to clarify here that our strategy in integrating correntropy matrices into the Bregman matrix divergence can be simply adapted to measure the discrepancy of two marginal distributions $p_1(x)$ and $p_2(x)$ with $D_{\varphi,B}(C_x^1\|C_x^2)$, in which $x\in \mathbb{R}^p$ and $C_x^1, C_x^2 \in \mathcal{S}^p_{+}$ refer to the correntropy matrix estimated from $p_1(x)$ and $p_2(x)$, respectively. Specifically, we have:
\begin{eqnarray}
D_{vN}(C_x^1\|C_x^2) = \Tr(C_x^1 \log C_x^1 - C_x^1 \log C_x^2 - C_x^1 + C_x^2),
\end{eqnarray}
and
\begin{eqnarray}
D_{\ell D}(C_x^1\|C_x^2) = \Tr((C_x^2)^{-1}C_x^1) + \log_2\frac{|C_x^2|}{|C_x^1|} - p.
\end{eqnarray}.

We can also achieve symmetry by taking the form:
\begin{eqnarray}
D_{vN}(C_x^1:C_x^2) = D_{vN}(C_x^1\|C_x^2) + D_{vN}(C_x^2\|C_x^1),
\end{eqnarray}
and
\begin{eqnarray}
D_{\ell D}(C_x^1:C_x^2) = D_{\ell D}(C_x^1\|C_x^2) + D_{\ell D}(C_x^2\|C_x^1).
\end{eqnarray}

Obviously, if one replaces correntropy matrix $C$ with covariance matrix $\Sigma$, the resulting LogDet divergence $D_{\ell D}(\Sigma_x^1\|\Sigma_x^2)$ reduces to the KL divergence for Gaussian data with $\mu_x^1=\mu_x^2$.

\subsection{Comparison with MMD in Two-Sample Test}

\begin{figure}[H]
\centering
\begin{subfigure}[]{0.45\textwidth}
\includegraphics[width=\textwidth]{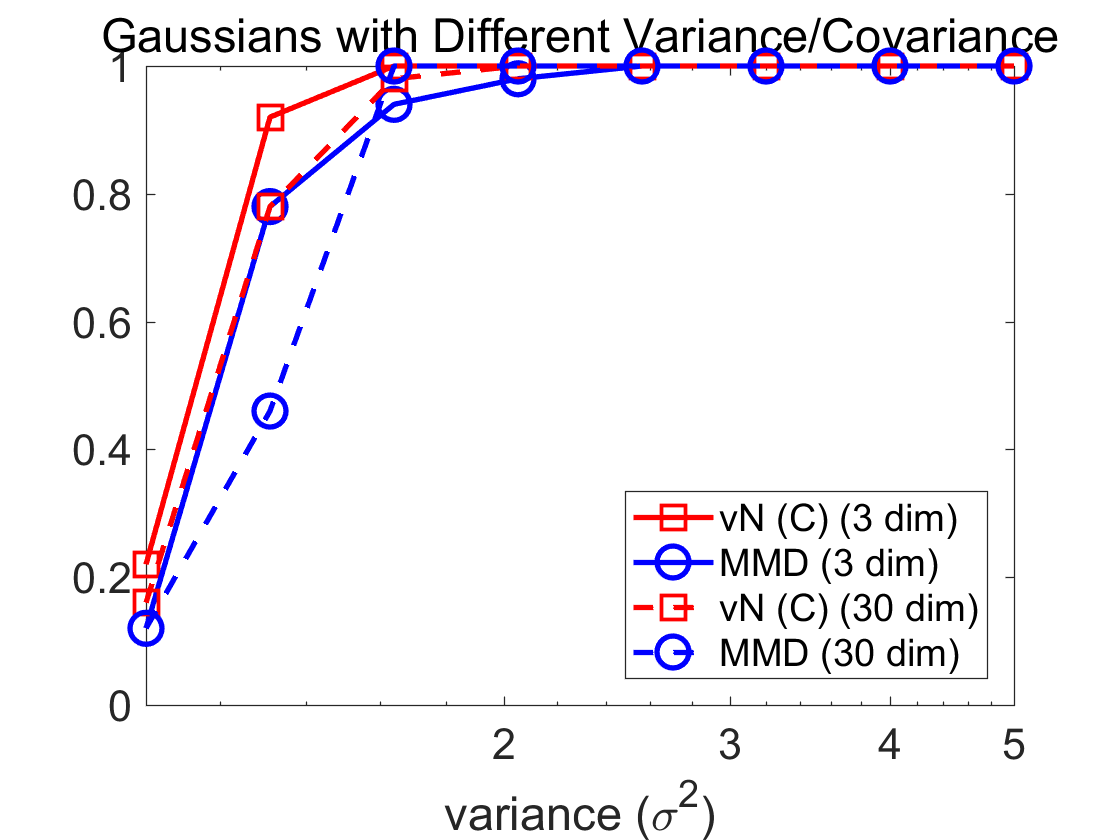}
\end{subfigure}\\
\begin{subfigure}[=]{0.45\textwidth}
\includegraphics[width=\textwidth]{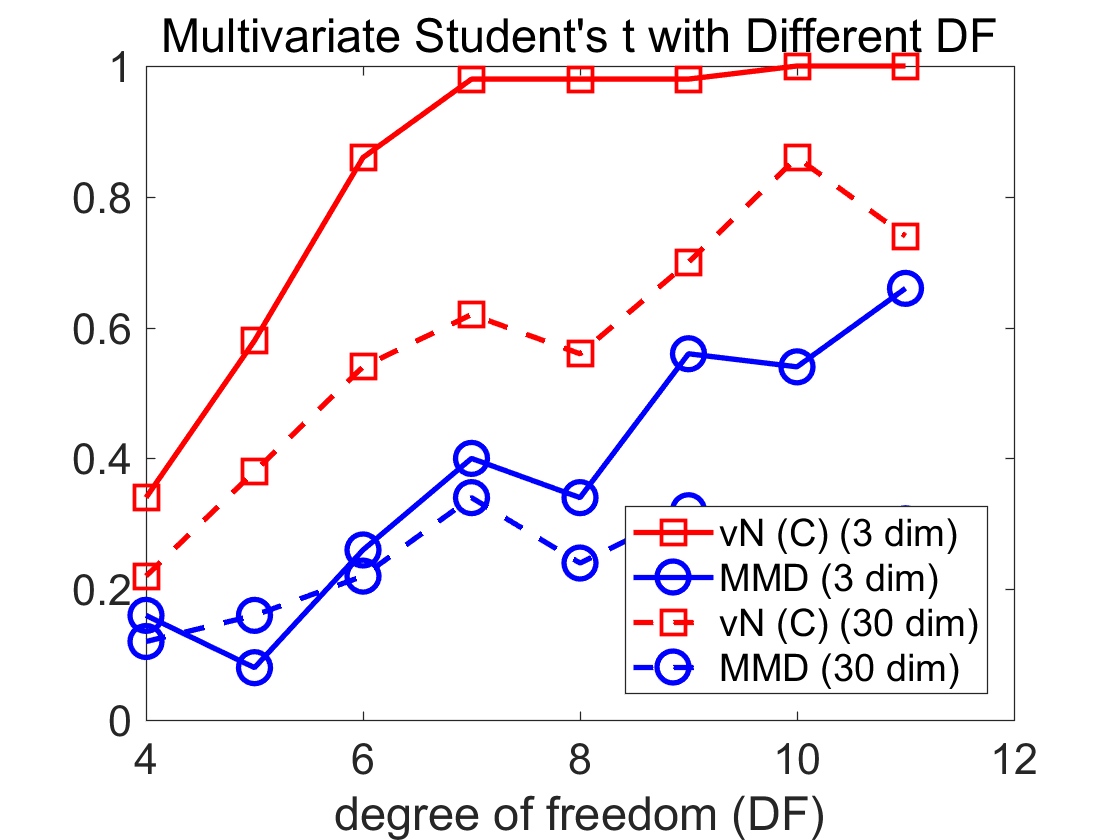}
\end{subfigure}
\caption{The power (i.e., $1$ minus Type-II error) of our von Neumann divergence (with correntropy matrix) and MMD in two-sample test. The sample dimensionality is indicated in the legend.}
\label{fig:power_test_MMD}
\end{figure}

We compare the discriminative power of our $D_{vN}$ and the classical maximum mean discrepancy (MMD) in a two-sample test on synthetic data. In the first simulation, our aim is to distinguish two isotropic Gaussian distributions which have the same mean but different covariance matrix (the first one is of covariance matrix $\mathcal{N}(0,\mathbf{I})$ whereas the second one is with $\mathcal{N}(0,\sigma^2\mathbf{I})$, where $\sigma^2$ is logarithmically distributed in the range $[1.05,5]$). In the second simulation, our aim is to distinguish two multivariate student's \textit{t}-distributions which have the same location but different shape matrix (the first one has a degree of freedom $3$, whereas the second one has degrees of freedom $\nu$, where $\nu = 4,5,\cdots,11$). Fig.~\ref{fig:power_test_MMD} suggests that our divergence is much more powerful than MMD for non-Gaussian distributions in high-dimensional space.

\subsection{Computational Complexity of Bregman-Correntropy Divergence}

Given two groups of observations, each has $n$ samples with dimension $d$, the computational cost of state-of-the-art divergence estimators relying on probability distribution estimation is always nearly $\mathcal{O}(n^3)$~\cite{kandasamy2015nonparametric}.
The computational complexity of our method is $\mathcal{O}(nd^2 + d^3)$ when using covariance matrix, while $\mathcal{O}(n^2d^2 + d^3)$ when using correntropy matrix, where $\mathcal{O}(nd^2)$ (or $\mathcal{O}(n^2d^2)$) is used for computing covariance (or correntropy) matrix and $\mathcal{O}(d^3)$ for computing eigenvalue decomposition in case of matrix logarithm (Eq.~(\ref{eq:vNRelativeEntropy})).

One should note that, correntropy complexity can be reduced to $\mathcal{O}(nd^2)$ (same to sample covariance)~\cite{li2020fast}. Our approach is advantageous when number of samples grows, which is common in current machine learning applications. When, on the other case, $d$ grows, although we suffer from high computational burden, the probability distribution estimation for other divergence estimators becomes inaccurate due to curse of dimensionality.

\subsection{Detect ``Salient” Pixels for Image Recognition}

\begin{figure}[ht]
\centering
\begin{subfigure}[]{0.5\textwidth}
\includegraphics[width=\textwidth]{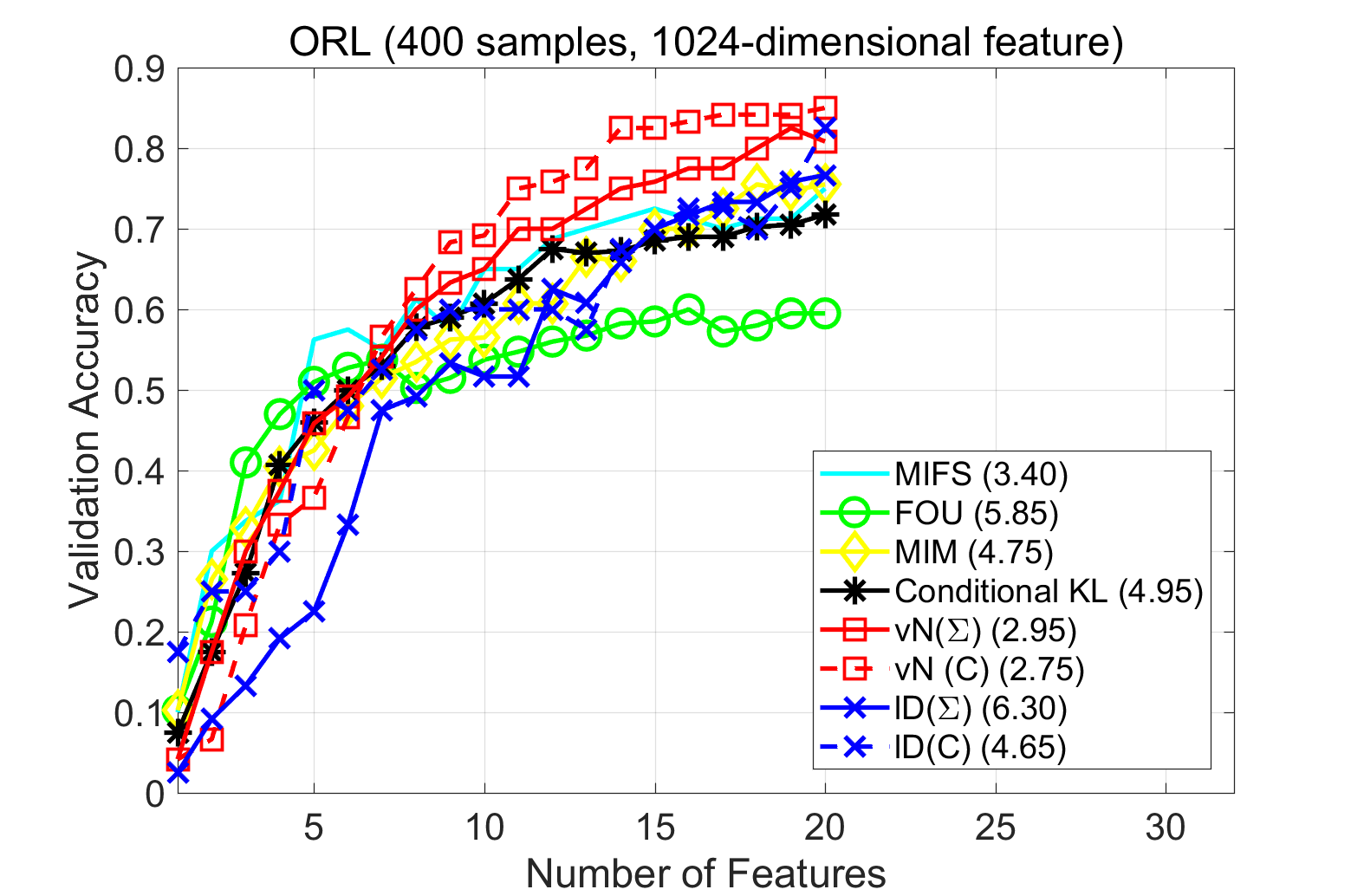}
\end{subfigure}\\
\begin{subfigure}[]{0.4\textwidth}
\includegraphics[width=\textwidth]{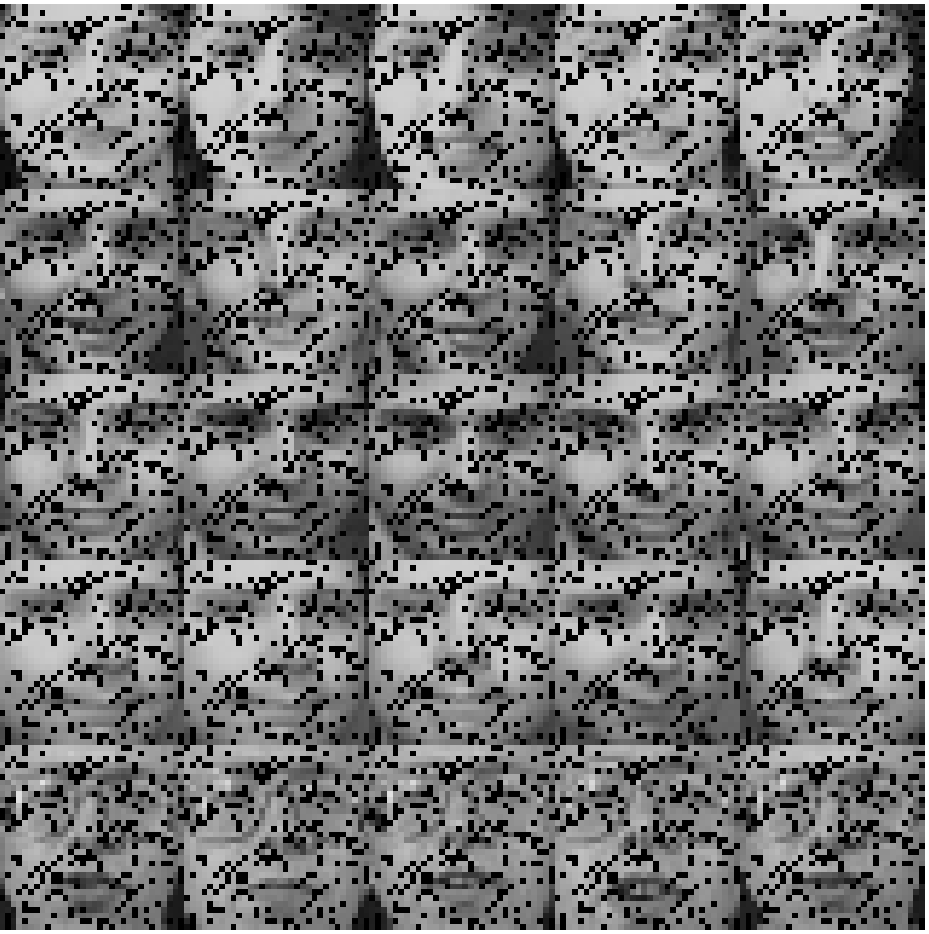}
\end{subfigure}
\caption{(a) demonstrates the validation accuracy on ORL face image data set (also known as AT\&T Database of Faces). The number of samples and the feature dimensionality for each data set are listed in the title. The number next to each method's name, in the legend, indicates the average rank of the method's performance in that data set. (b) visualizes the selected $200$ pixels (as black points) using vN ($C$) in randomly selected face images.}
\label{fig:ORL}
\end{figure}

The feature selection method developed in Section~\ref{sec:feature} can be used to detect ``salient” or significant pixels for image recognition as well. We verify this claim with the benchmark ORL face recognition data set. As can be seen in Fig.~\ref{fig:ORL}, our von Neumann divergence with correntropy matrix always has the highest validation accuracy. By incorporating higher order information, centered correntropy matrix (i.e., $C$) performs better than its covariance based counterpart (i.e., $\Sigma$). Moreover, our method always selects pixels that are lying on edges such as eyelids, mouth corners and nasal bones, which makes our result becomes interpretable.

\subsection{The Automatic Differentiability of Bregman-Correntropy Divergence}
We finally want to underline that our Bregman-Correntropy divergence is differentiable and can be used as loss functions to train neural networks. To exemplify this property, we consider here the training of a deep generative autoencoder.

\begin{figure}[H]
\centering
\includegraphics[width=0.45\textwidth]{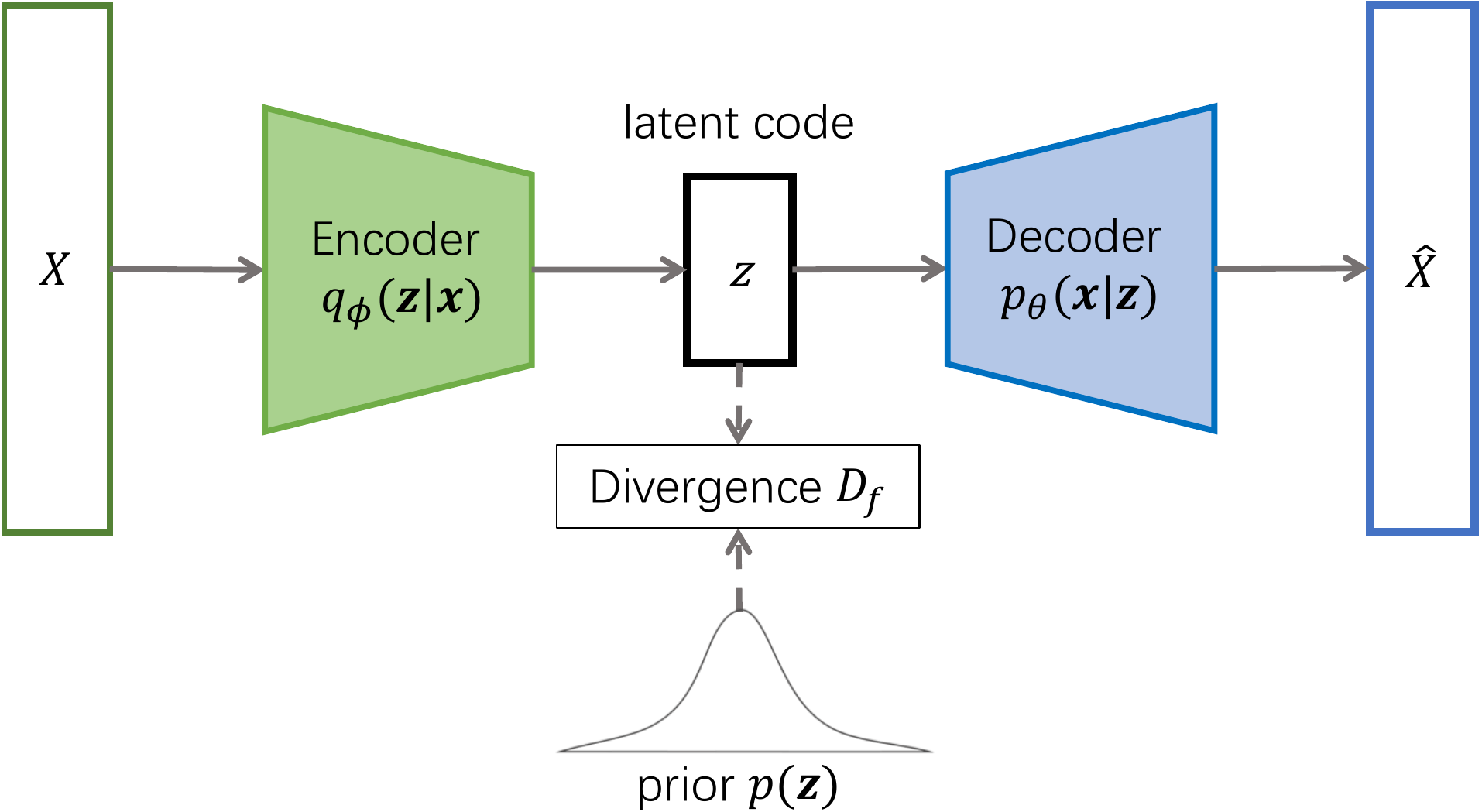}
\caption{The framework of a deep generative autoencoder which is specified by the encoder, decoder, and the prior distribution on the latent (representation/code) space. The encoder $E_\phi$ maps the input to the representation space, while the decoder $D_\theta$ reconstructs the original input from the representation. The latent code is encouraged to match a prior distribution $p(z)$ by a divergence $D_f$.}
\label{fig:VAE_illustration}
\end{figure}

The adversarial autoencoders (AAEs)~\cite{makhzani2015adversarial} have the architecture that inspired our model the most. Specifically, AAEs turn a standard autoencoder (rather than the variational autoencoder~\cite{kingma2013auto}) into a generative model by imposing a prior distribution $p(z)$ on the latent variables by penalizing some statistical divergence $D_f$ between $p(z)$ and $q_\phi(z)$ (see Fig.~\ref{fig:VAE_illustration} for an illustration). Using the negative log-likelihood as reconstruction loss, the AAE objective can be written as:
\begin{eqnarray}\label{eq:loss_AAE}
    \mathcal{L}_{AAE}(\theta,\phi)=\mathbb{E}_{\hat{p}(x)}\left[\mathbb{E}_{q_\phi(z|x)}\left[-\log{p_\theta(x|z)}\right]\right] \nonumber \\
    +\lambda D_f\left(q_\phi(z)\|p(z)\right).
\end{eqnarray}

During implementation, the encoder and decoder are taken to be deterministic, and the negative log-likelihood in Eq.~(\ref{eq:loss_AAE}) is replaced with the standard autoencoder reconstruction loss:
\begin{equation}
    \mathbb{E}_{\hat{p}(x)}\left[\|x-D_\theta \left(E_\phi(x)\right)\|_2^2\right].
\end{equation}

AAEs use an additional generative adversarial network (GAN)~\cite{goodfellow2014generative} to measure the divergence from $p(z)$ to $q_\phi(z)$. The advantage of implementing a specific regularizer $D_f$ (e.g., a GAN) is that any $p(z)$ we can sample from, can be matched~\cite{tschannen2018recent}.

Unlike AAEs, we define a new form of regularization from first principles, which allows us to train a competing method with much fewer trainable parameters. Specifically, we explicitly replace $D_f$ based on GAN with our Bregman-Correntropy divergence $D_{\varphi,B}$, and train a deterministic autoencoder with the following objective:
\begin{eqnarray}
    \mathcal{L}_{ours}(\theta,\phi)=\mathbb{E}_{\hat{p}(x)}\left[\|x-D_\theta \left(E_\phi(x)\right)\|_2^2\right] \nonumber \\
    +\lambda D_{\varphi,B}(C_{q_{\phi(z)}}\|C_{p(z)}),
\end{eqnarray}
where $C_{q_\phi(z)}$ and $C_{p(z)}$ denote, respectively, the correntropy matrices evaluated from $q_\phi(z)$ and $p(z)$.

We set the network architecture as $784-1000-1000-2-1000-1000-784$ and $\lambda=1$. We use kernel size $\sigma=10$ to estimate the centered correntropy. The optimizer is Adam with learning rate $1e-3$.

\begin{figure}[ht]
\centering
\begin{subfigure}[]{0.23\textwidth}
\includegraphics[width=\textwidth]{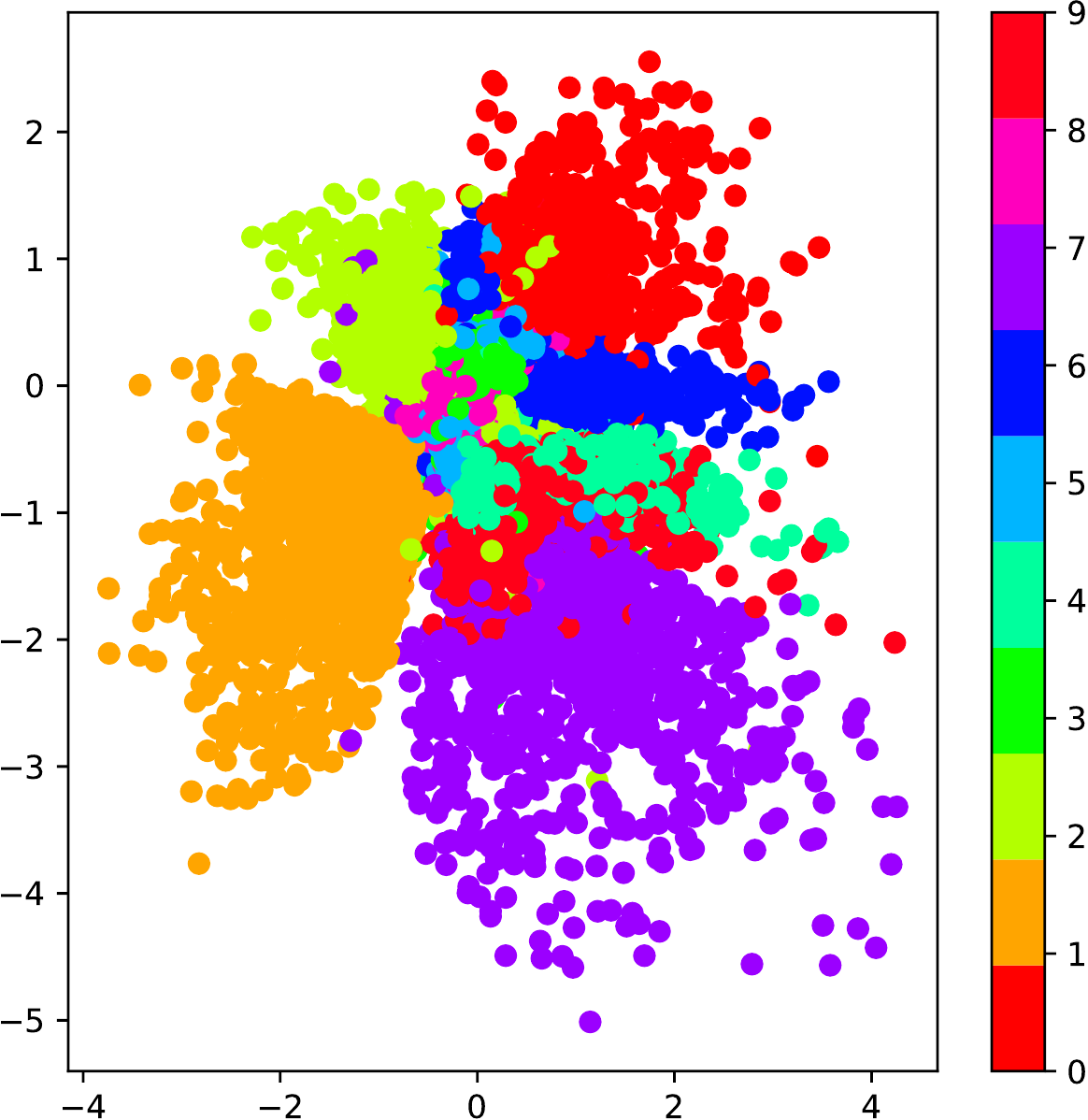}
\end{subfigure}
\begin{subfigure}[=]{0.23\textwidth}
\includegraphics[width=\textwidth]{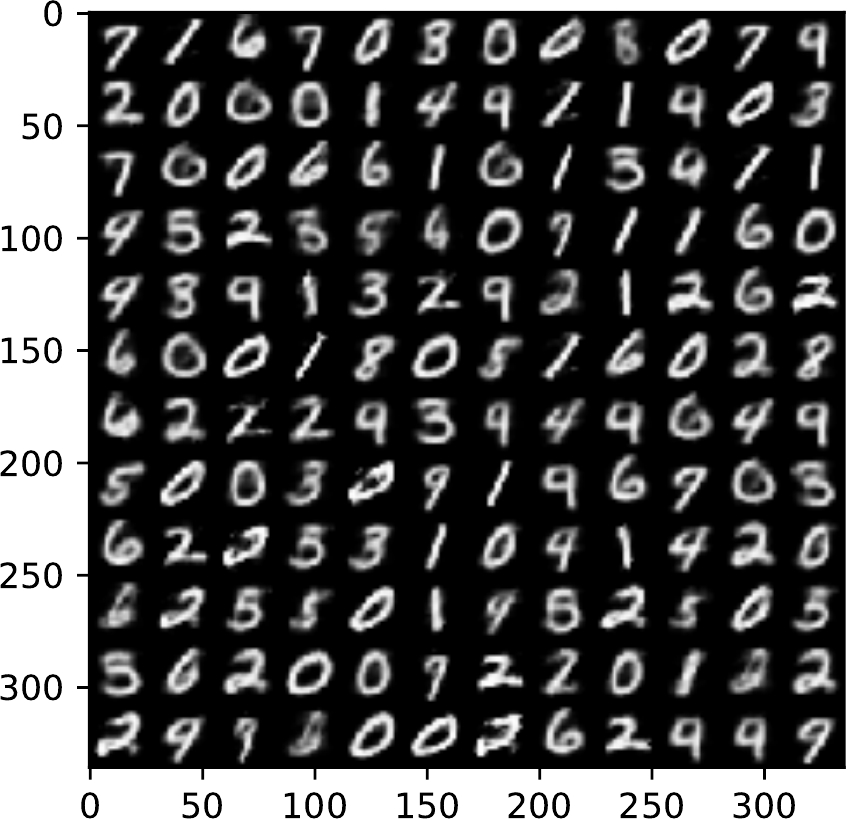}
\end{subfigure} \\
\begin{subfigure}[]{0.23\textwidth}
\includegraphics[width=\textwidth]{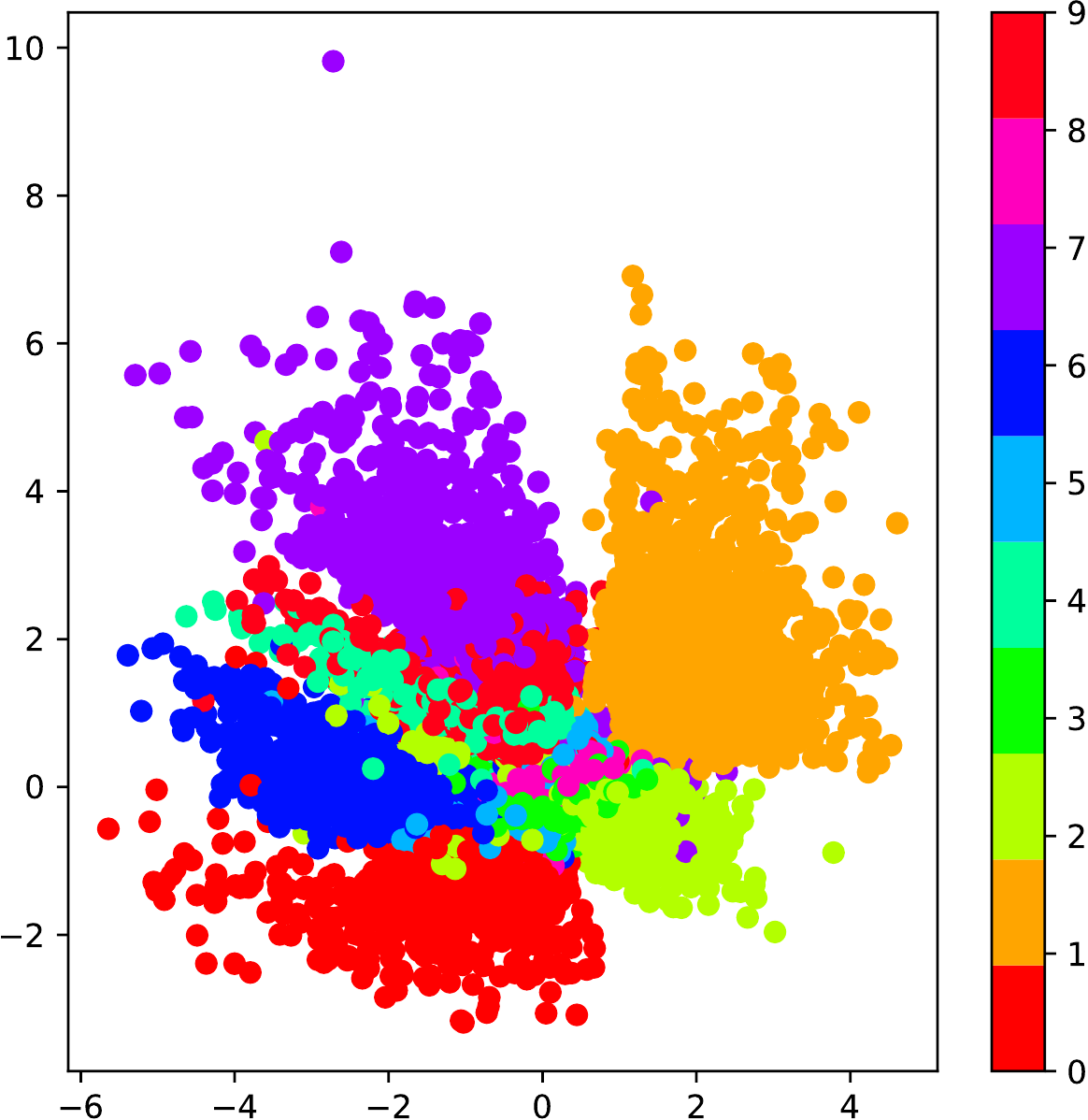}
\end{subfigure}
\begin{subfigure}[=]{0.23\textwidth}
\includegraphics[width=\textwidth]{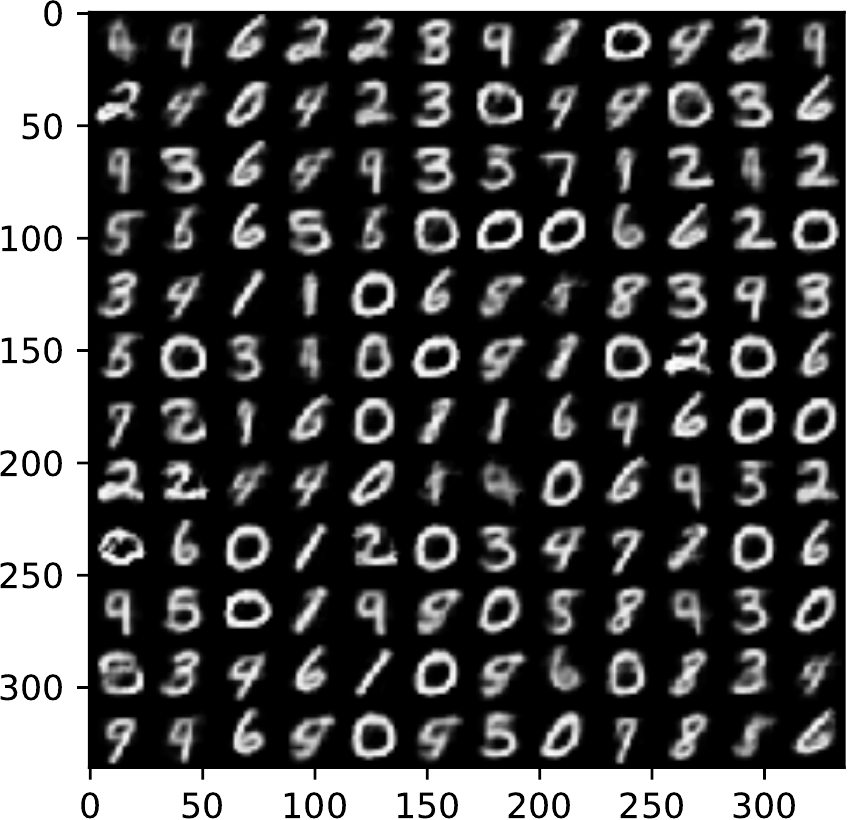}
\end{subfigure}
\caption{The latent code distribution and the generated images by sampling from a prior distribution $p(z)$. In the first row, $p(z)$ is a Gaussian; in the second row, $p(z)$ is a Laplacian.}
\label{fig:latent_images}
\end{figure}

We show, in Fig.~\ref{fig:latent_images}, the latent code distribution and the generated images by sampling from a predefined prior distribution $p(z)$. As can be seen,$D_{\varphi,B}$ enables different kinds of $p(z)$ (not limited to an isotropic Gaussian as in VAE) and our loss provides a novel alternative to train deep generative autoencoders.

\end{document}